# Evolution of AI in Education: Agentic Workflows

Firuz Kamalov[1]*, David Santandreu Calonge[2]*, Linda Smail[3], Dilshod Azizov[4], Dimple R. Thadani[5], Theresa Kwong[6], Amara Atif[7]

1 Department of Electrical Engineering, Canadian University Dubai, Dubai, United Arab Emirates, 2 Centre for Teaching and Learning, Mohamed bin Zayed University of Artificial Intelligence, Abu Dhabi, United Arab Emirates, 3 College of Interdisciplinary Studies, Zayed University, Dubai, United Arab Emirates, 4 Mohamed bin Zayed University of Artificial Intelligence, Abu Dhabi, United Arab Emirates, 5 Department of Management, Marketing, and Information Systems, Hong Kong Baptist University, Hong Kong SAR, 6 Centre for Holistic Teaching and Learning, Hong Kong Baptist University, Hong Kong SAR, 7 School of Computer Science, University of Technology Sydney, Sydney, Australia

*firuz@cud.ac.ae, david.santandreu@mbzuai.ac.ae

**Abstract**

The primary goal of this study is to analyze agentic workflows in education according to the proposed four major technological paradigms: reflection, planning, tool use, and multi-agent collaboration. We critically examine the role of AI agents in education through these key design paradigms, exploring their advantages, applications, and challenges. Second, to illustrate the practical potential of agentic systems, we present a proof-of-concept application: a multi-agent framework for automated essay scoring. Preliminary results suggest this agentic approach may offer improved consistency compared to stand-alone LLMs. Our findings highlight the transformative potential of AI agents in educational settings while underscoring the need for further research into their interpretability and trustworthiness.

Keywords: AI Agents; Education; Artificial Intelligence; Automated Essay Scoring; LLM; GPT; Sustainability

# 1 Introduction

Artificial intelligence (AI) has rapidly evolved in recent years, increasingly permeating various aspects of modern society, including education. The emergence of large language models (LLMs) has particularly expanded AI's capabilities, creating opportunities to develop educational tools that not only enhance learning outcomes but also raise awareness of sustainable practices. Early applications in education largely relied on foundational models such as GPT or their fine-tuned variants (ChatGPT). While these models facilitated notable advancements in instruction and student support, they also operated under inherent constraints—limited adaptability, lack of nuanced reasoning, and reliance on static training data—that underscore the importance of ongoing research into sustainable, responsible AI innovations.

A new generation of AI technologies seeks to overcome the limitations of the foundational models through the development of AI agents that function autonomously. AI agents utilize reflection, planning, and a wide range of tools to perform tasks with minimal human intervention. At the core of most AI agents lie sophisticated LLMs; however, their agentic layer distinguishes them from traditional systems by enabling more complex capabilities and autonomy. Specifically, agentic design leverages real-time information retrieval and dynamic task decomposition, reducing the reliance on dated or incomplete training data. As a result, AI agents are able to outperform stand-alone LLMs on numerous benchmark evaluations (Hong et al., 2024; Wu et al., 2024), demonstrating the considerable promise of agentic workflows in education (Guo et al., 2024).

Although the implementation of AI agents within educational settings has begun to be investigated (Yusuf et al., 2025; Dai et al., 2024), their integration into educational settings requires careful examination and strategic implementation. Educational tasks require robust, trustworthy, and interpretable systems. Further research is needed to determine how AI-driven agents influence different learning environments, student populations, and pedagogical methods.

Historic work on agents in AI in education, as represented by Johnson et al. (2000) and Kim et al. (2006), primarily focused on embodied pedagogical agents with predefined behaviors designed to enhance learning through social presence and scripted interactions within constrained educational scenarios. These early agents functioned as content-delivery mechanisms with limited autonomy, emphasizing visual representation and basic dialogue systems to simulate human-like tutoring within narrow domains. In contrast, contemporary agentic-AI in education, as reflected in recent research including Zhang et al. (2024 a&b) and Azevedo et al. (2022), represents a paradigm shift toward systems exhibiting advanced capabilities including reflection, strategic planning, tool utilization, and multi-agent collaboration. Modern agentic-AI moves beyond static single-purpose agents to create dynamic systems that autonomously navigate complex educational workflows, adapt to learner needs in real-time, and engage in sophisticated multi-agent coordination—transforming from simple instructional aides to proactive partners in the learning process with significantly greater autonomy, adaptability, and capacity for managing educational processes rather than merely delivering content.

The primary goal of this paper is to critically analyze the existing AI agentic paradigms within the context of education. We propose a technological taxonomy based on four major design paradigms: 1) reflection, 2) planning, 3) tool use, and 4) multi-agent collaboration. These paradigms are “crucial for boosting LLM productivity and enhancing performance” (Singh et al., 2024, p.1). We first describe each paradigm individually, discussing their conceptual foundations and respective advantages, practical applications, and associated challenges within educational contexts. While current research on Agentic AI in education often focuses on singular capabilities like automated tutoring or content generation, our work provides a novel, structured framework for analyzing and designing these systems.  Our work explicitly shifts the focus to the educational implications, advantages, and challenges of each paradigm in practice The present analysis provides a comprehensive understanding of how these paradigms can effectively leverage AI agents to overcome the limitations of conventional large language models.

Furthermore, a significant gap exists in synthesizing the new wave of LLM-based agentic paradigms specifically through an educational lens. Our work makes an original contribution by proposing and applying a unified analytical framework—based on reflection, planning, tool use, and multi-agent collaboration—to the specific context of education. Unlike existing technical reviews, our work does not merely describe these paradigms; it critically examines their pedagogical implications, practical applications, and inherent challenges within learning environments.

Second, to illustrate the potential of an agentic approach in education, we present a proof-of-concept application: a multi-agent system designed for automated essay scoring (MASS). The results show that our proposed agentic framework achieves increased consistency and reliability in evaluating written work compared to stand-alone LLMs (https://github.com/AzizovDilshod/Multi-agent-System-for-Essay-Assessment/tree/main).

Although several prior reviews have discussed agentic architectures, performance benchmarks, and capability taxonomies (e.g., Wang et al., 2023a; Xi et al., 2023; Singh et al., 2024), our study advances a qualitatively different contribution. To the best of our knowledge, this paper presents the first systematic effort to categorize agentic methods and architectures especially with the intent to map their functional roles onto pedagogical affordances for education. The shift from cataloguing technical capability to grouping methods by educational function and consequence is intended to make agentic research actionable for instructional design, evaluation, and policy in educational settings.

The rest of the manuscript is organized as follows. In Section 2, we provide a conceptual overview of AI agents and agentic workflows. Crucially, we propose our categorization of agentic workflows in education in terms of four major paradigms (Figure 2.4). Section 3, describes the structured literature review methodology used to analyze the role and potential of AI agentic workflows in education. Our findings are synthesized narratively within Section 4, dedicating a subsection to each paradigm. Section 5 details our proposed multi-agent system for automated essay scoring, outlining its architecture, implementation, and empirical results. Finally, Section 6 concludes with a discussion on the limitations of existing approaches and directions for future work in advancing agentic solutions for education.

# 2 Background: AI Agents and Agentic Workflows

An AI agent can be defined as a system that utilizes an LLM as its core reasoning engine to determine the control flow of an application, interacting with its environment to achieve specific goals (Wang et al., 2023a; Xi et al., 2023). An agentic workflow refers to the structured process through which such autonomous or semi-autonomous agents operate. Unlike traditional LLMs, which are often constrained by the static knowledge embedded during training, AI agents are designed for dynamic interaction. They can actively retrieve real-time information, utilize various tools (e.g., calculators, databases, web search), and employ reasoning strategies to execute complex tasks (Fig. 2.1). This capacity for autonomous action with limited human intervention makes agentic workflows a powerful paradigm for building sophisticated LLM-based applications.

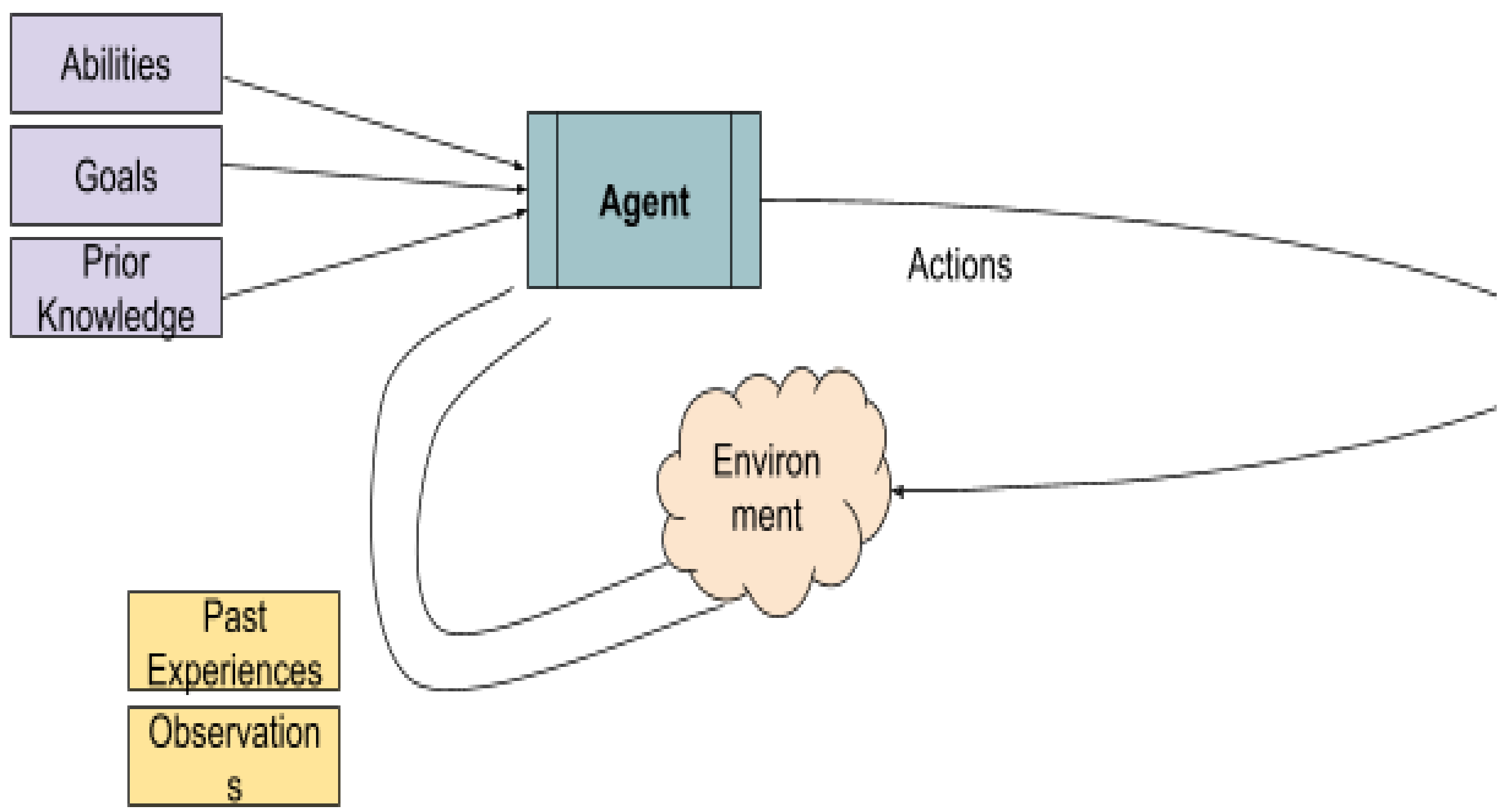

**Fig. 2.1** In the basic agentic workflow, the agent interacts with its environment based on its goals and abilities.

The shift towards agentic systems is partly driven by the evolving landscape of AI development. While past progress in LLMs was fueled by scaling data and computation, limitations in data availability and computational resources are encouraging a focus on enhancing inference-time performance (Zeng et al., 2024). Techniques such as chain-of-thought prompting, self-consistency, ReAct (Reasoning and Acting), and reflection, which are often incorporated into agentic workflows, have been significant drivers of recent performance improvements (Shinn et al., 2023).

Evidence suggests that agentic systems can outperform their stand-alone LLM counterparts across various domains. As traditional benchmarks are increasingly mastered, more rigorous evaluation frameworks like 'Humanity's Last Exam' are emerging to test the upper limits of general AI capabilities (Phan et al., 2025). The complexity of these new challenges further underscores the necessity of agentic workflows that go beyond simple inference to include multi-step reasoning and tool use. For example, benchmark results on tasks like HumanEval show significant performance gains when agentic frameworks are built upon base models like GPT-3.5 and GPT-4 (Fig. 2.2) (Ng, 2024). Similar improvements have been observed in coding benchmarks like TransCoder and MBPP. In education, preliminary studies indicate that agentic systems can enhance tasks such as automated scoring (Lee et al., 2024; Guo et al., 2024), demonstrating the potential of these approaches within the educational domain.

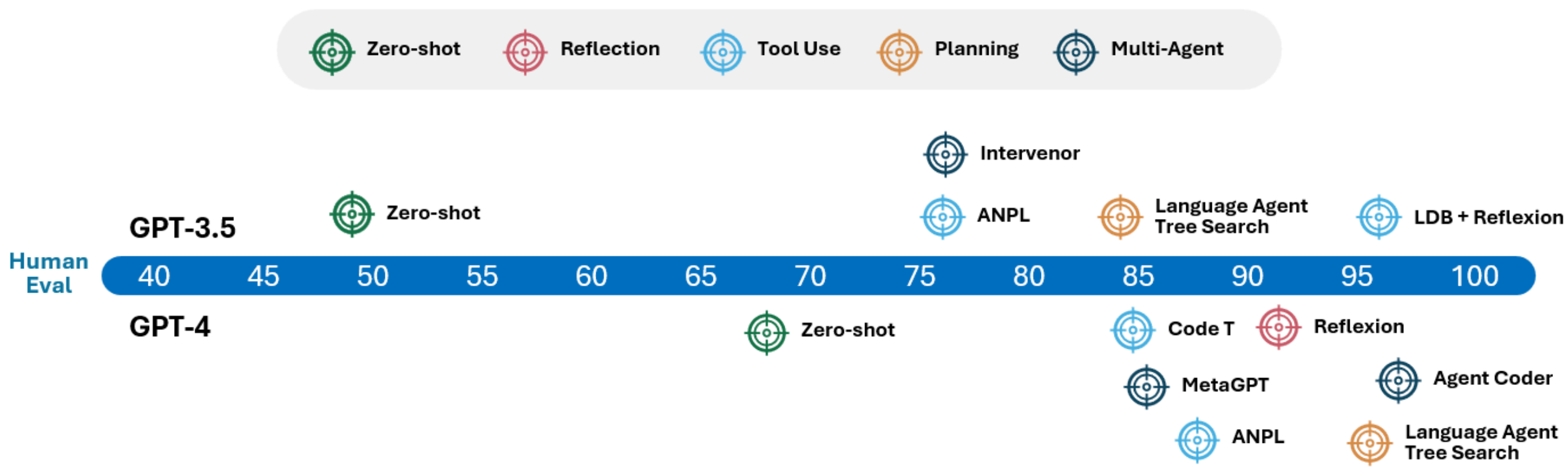

**Fig. 2.2** Performance of GPT 3.5 and GPT-4 (zero-shot) on HumanEval along with agentic frameworks built on top of GPT (Ng, 2024).

Conceptually, AI agents are forming a distinct "application" layer in the evolving AI technology stack (Fig. 2.3). While foundational models, cloud infrastructure, and hardware remain crucial, the agent layer focuses on orchestrating these components to execute complex, goal-oriented tasks, abstracting underlying complexity and enabling more seamless user interaction.

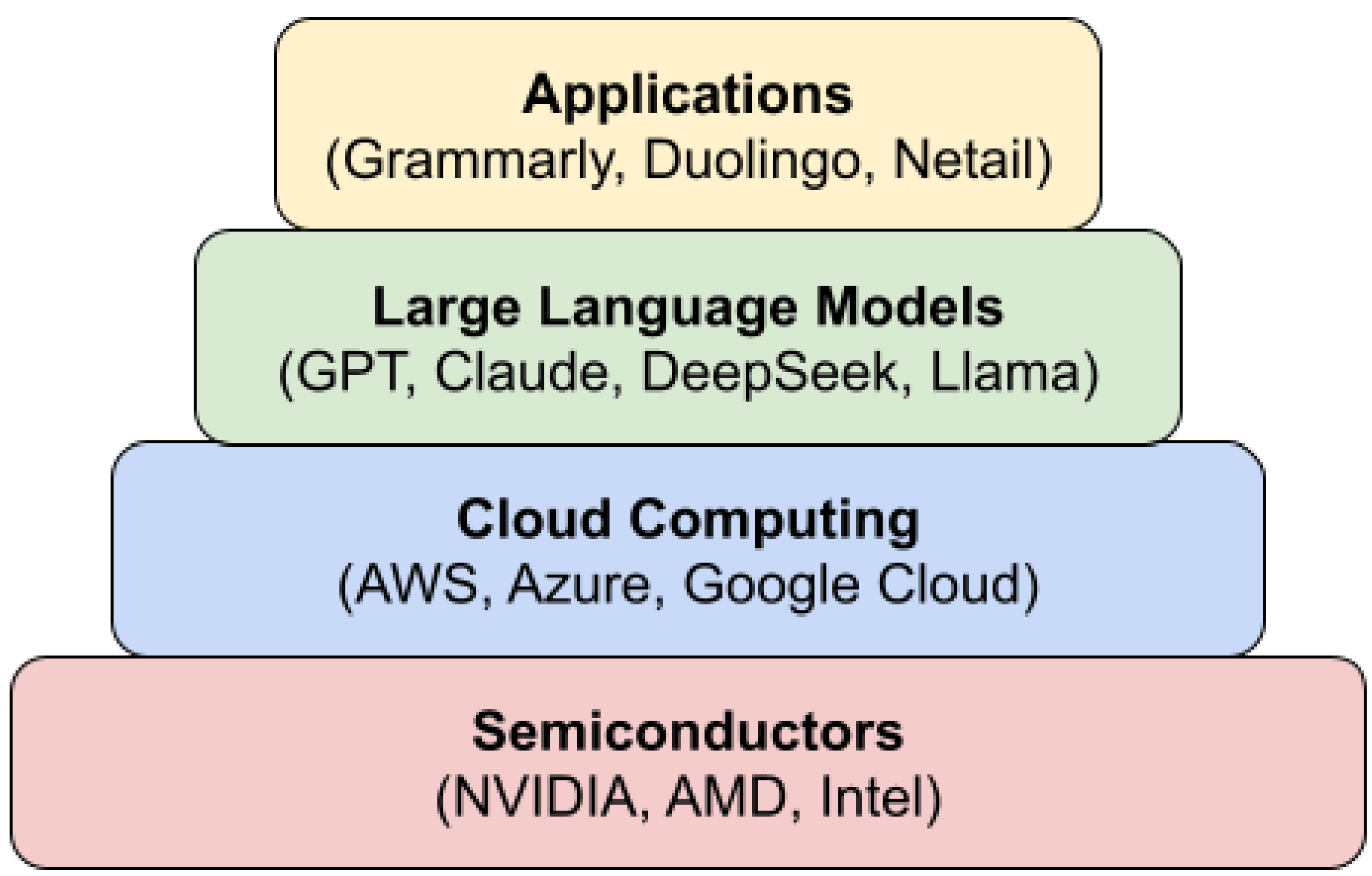

**Fig. 2.3** The new AI stack featuring the emerging layer of Applications.

## 2.1 Agentic Workflow

A general agentic workflow often involves several key phases, though specific implementations vary:

1. Planning: The agent receives a goal, decomposes it into smaller, manageable sub-tasks, and sequences these tasks.

2. Information Retrieval/Tool Use: The agent identifies information needs for each sub-task and utilizes available tools (e.g., web search, databases, APIs, other agents) to gather necessary data or perform specific actions.
3. Task Execution & Refinement: The agent executes the planned tasks, monitors progress, potentially reflects on outcomes or feedback, and dynamically adjusts the plan or generates new tasks as needed until the overall goal is achieved.

The degree of "agenticness" can vary widely, from a simple system using a single tool to complex multi-agent collaborations. As the technology matures, agents are expected to become increasingly autonomous and capable (e.g., OpenAI's advancements, DeepSeek-R1).

Four major paradigms underpin many current agentic approaches (Ng, 2024; Singh et al., 2024), as shown in Figure 2.4:

1. Reflection: The agent analyzes its past actions or outputs to identify errors or areas for improvement and refine its future behavior.
2. Planning: The agent explicitly creates and follows a sequence of steps or sub-goals to achieve a complex objective.
3. Tool Use: The agent leverages external resources or functions (e.g., calculators, code execution, web search, APIs) to augment its capabilities.
4. Multi-agent Collaboration: Multiple specialized agents work together, communicating and coordinating their actions to achieve a common goal.

These paradigms, which form the basis of our review in Section 4, represent distinct but often complementary strategies for building effective AI agents.

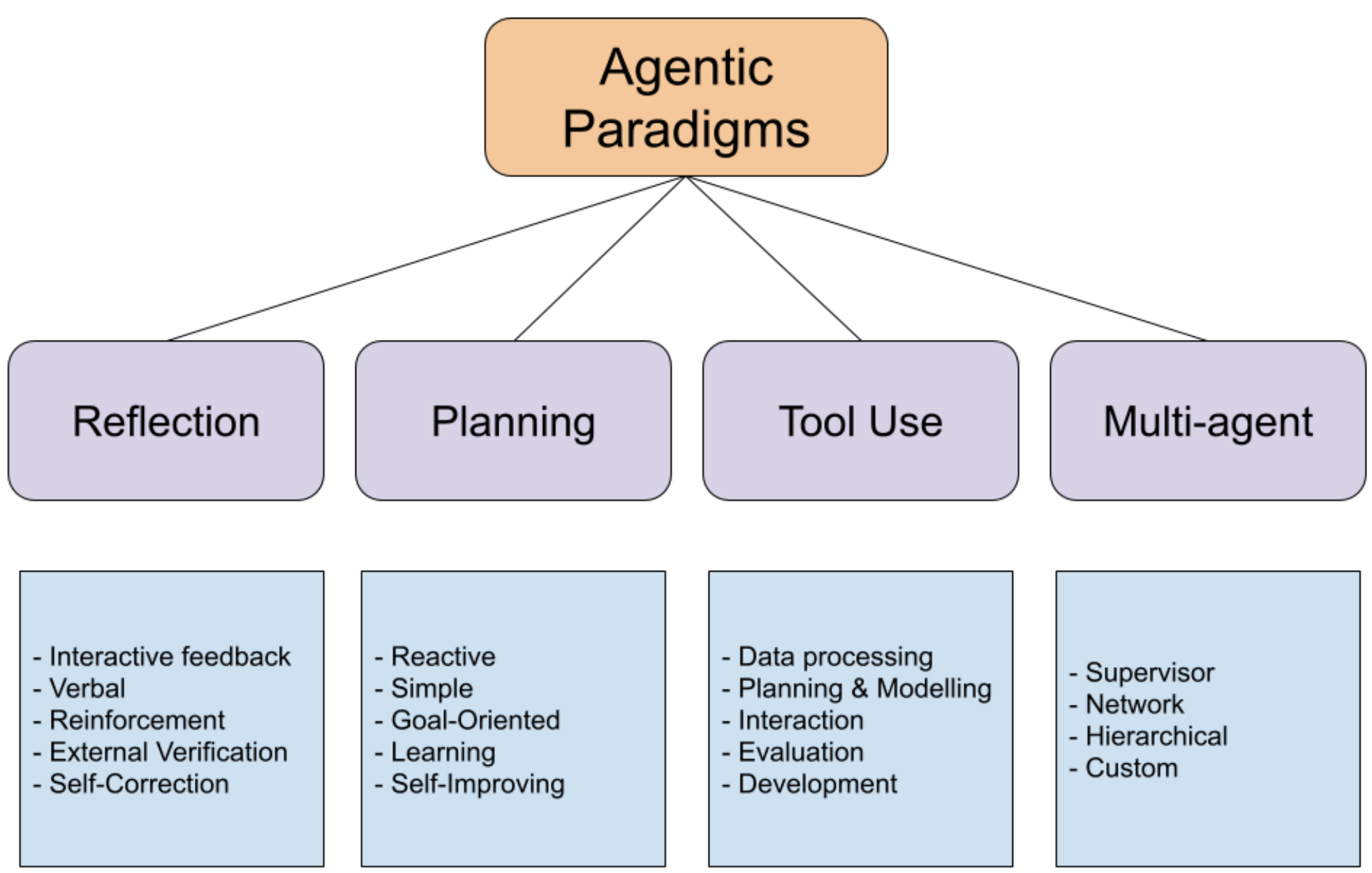

**Fig. 2.4** Description of the four major agentic paradigms

## 2.2 AI Agent Platforms

Several platforms facilitate the development and implementation of agentic systems, such as AutoGen, CrewAI, LangGraph, MetaGPT, and Phidata (Table 2.1). These frameworks provide varying levels of structure, flexibility, and support for different agent architectures, enabling researchers and practitioners (including educators) to build custom agent-based applications.

**Table 2.1**. Platforms for building and implementing agentic systems.

| **Framework** | **Key Features** | **Pros** | **Cons** |
|---|---|---|---|
| AutoGen (Wu et al., 2024) | Open-source, flexible design, active community | - Highly customizable - Integrates with various tools and human feedback | - Steep learning curve - Potentially less structured than other frameworks |
| MetaGPT (Hong et al., 2024) | Standardized prompt sequences, a rich library of predefined agents | - Excels in orchestrating complex agent interactions - Reduces custom coding | - Heavy reliance on asyncio - Less generalizable agent roles |
| CrewAI (CrewAI, 2024) | Production-oriented, structured role-based delegation | - Clean, production-ready design - Facilitates predictable, reliable outcomes | - Limits on agent re-delegation - Collects anonymized usage data, which may raise privacy concerns |
| LangGraph (LangChain, 2024) | Graph-based multi-actor workflow, fine-grained state control | - Clear, scalable approach to multi-agent interactions - Encourages task specialization | - Requires familiarity with graph theory- May be less suitable for broad, highly interconnected tasks |

# 3 Methodology

This paper employs a structured literature review methodology, following the PRISMA 2020 guidelines to ensure transparency and reproducibility. The primary objective is to synthesize current research and identify key trends, applications, benefits, and challenges associated with four major agentic paradigms: reflection, planning, tool use, and multi-agent collaboration, as identified in recent surveys and framework discussions (e.g., Ng, 2024; Singh et al., 2024; Xi et al., 2023).

## 3.1 Literature Search Strategy

A comprehensive search was conducted across four major academic databases: Google Scholar, Scopus, Web of Science, and the ACM Digital Library, yielding 657 records. An additional 231 records were identified through supplementary methods, including backward and forward reference list checking, grey literature searches, and expert consultation. In total**,** 888 records were retrieved.

Search terms included combinations and variations of primary keywords such as *"AI agents," "agentic workflows," "large language models,"* and *"education,"* combined with paradigm-specific terms like *"reflection," "planning," "tool use," "multi-agent systems," "autonomous agents," "agentic AI,"* and related concepts (e.g., *"ReAct," "Reflexion," "chain-of-thought," "CRITIC," "SELF-REFINE"*). Application-specific terms such as *"automated tutoring," "personalized learning," "educational simulation," "automated scoring,"* and *"adaptive learning"* were also used to capture relevant educational contexts.

## 3.2 Selection Criteria and Scope

All records were imported into a reference manager for deduplication, leaving 378 unique records. Titles and abstracts were independently screened by two reviewers against predefined inclusion criteria. At this stage, 94 records were excluded, resulting in 284 full-text articles for eligibility assessment.

During the full-text review, 191 articles were excluded for the following reasons:

1. Focused solely on foundational LLM capabilities without an agentic layer (*n = 75*).
2. Addressed AI applications in education unrelated to agentic workflows (*n = 67*).
3. Not available in English (*n = 45*).
4. Primarily commercial or promotional content lacking technical or conceptual depth (*n = 22*).

Following this process, 93 studies were included in the final synthesis. Discrepancies during screening and eligibility assessment were resolved through discussion or consultation with a third reviewer. The flow of studies through the review process is summarized in Figure 3.1.

Included studies consisted of peer-reviewed journal articles, conference proceedings, and highly cited arXiv. Framework documentation and conceptual pieces from reputable sources were also considered where they defined influential paradigms or architectures.

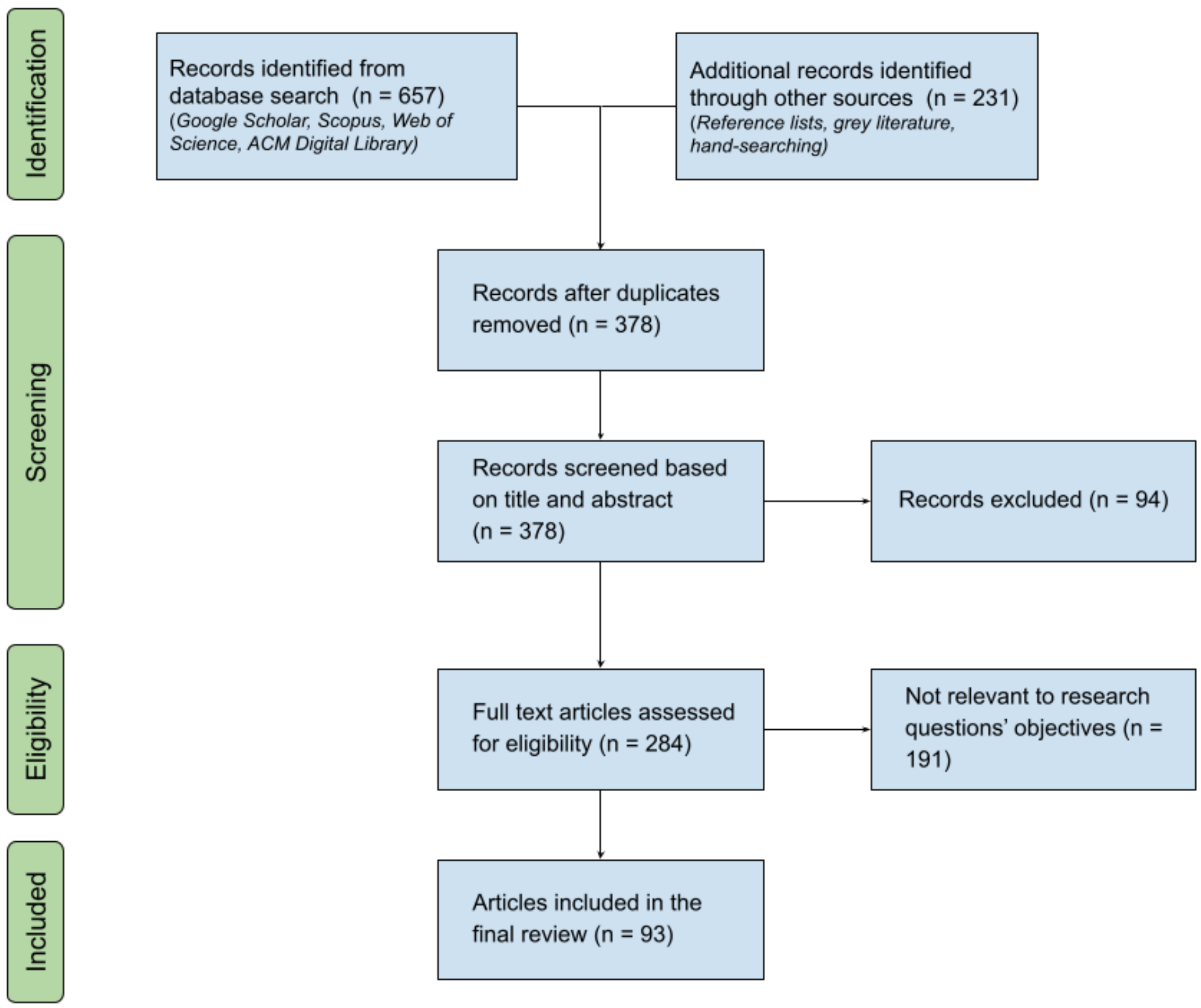

**Fig. 3.1** PRISMA flow chart

## 3.3 Study Quality and Risk-of-Bias Assessment

To evaluate the reliability and significance of the included studies, we conducted a formal methodological quality appraisal and risk-of-bias assessment for all 93 included records. Because the corpus contains empirical studies, conference papers, conceptual and framework papers, and highly cited preprints, we used an adapted mixed-methods appraisal approach based on the Mixed Methods Appraisal Tool critical appraisal checklists. The adapted instrument assessed each study on the following domains:

1. Study type and peer-review status.
2. Clarity of aims and research questions.
3. Appropriateness of methods.
4. Data and sampling.
5. Analysis and reporting.
6. Validity and reliability.
7. Conflict of interest / funding transparency.

Each domain was rated as Yes / Partial / No / Not applicable and studies were assigned an overall categorical rating of High, Moderate, or Low methodological quality according to predefined scoring rules. Two reviewers (FK, DSC) independently rated each study. Discrepancies were resolved by discussion and, where necessary, consultation with a third reviewer (LS). We summarize the appraisal results below (Table 3.2) and provide the full item-level ratings for each included study in Appendix A (supplementary material).

## 3.4 Analysis Framework

The selected literature was analyzed using a framework structured around the four core paradigms. For each paradigm, the analysis focused on extracting and synthesizing information related to:

1. Conceptual Foundations: Core principles, mechanisms, and key enabling techniques.
2. Educational Applications: Reported or proposed uses in education.
3. Advantages: Benefits highlighted compared to non-agentic or traditional methods.
4. Challenges and Future Directions: Reported limitations, implementation hurdles, ethical, and suggested research avenues.

## 3.5 Quality Appraisal

The methodological quality of the 93 included studies was heterogeneous and skewed toward conceptual and early empirical work. Based on the adapted appraisal described in Section 3.3, the aggregated overall ratings are shown in Table 3.2. Full item-level ratings are available from the authors on request.

Table 3.2. Summary of methodological quality of included studies.

| **Overall rating** | **Number of studies** | % of included studies |
|---|---|---|
| High | 13 | 14% |
| Moderate | 46 | 49% |
| Low | 34 | 37% |
| Total | 93 | 100% |

# 4 Agentic Paradigms in Education

As outlined in Section 2, agentic workflows can be broadly categorized into four key paradigms: reflection, planning, tool use, and multi-agent collaboration. This section reviews each paradigm, discussing its foundations, specific roles and applications in education, and associated challenges.

## 4.1 Reflection Systems

AI-driven Reflection systems are designed to enable agents to evaluate their performance, identify areas for improvement, and iteratively refine their approaches to achieve tasks more effectively. By continually monitoring and adjusting strategies, these systems embody principles of sustainable growth—where perpetual learning and responsible resource use guide progress. This paradigm aligns closely with foundational principles from the science of learning, particularly those emphasizing the importance of metacognition and active engagement in learning processes (Kosslyn, 2017). In educational contexts, reflection systems offer transformative potential not only by fostering adaptive learning experiences and enhancing teaching methodologies and student outcomes but also by promoting awareness of sustainable practices. Through iterative feedback loops, such systems can encourage learners and educators alike to consider long-term impacts, optimize resource utilization, and integrate environmental stewardship into the learning process.

These systems incorporate mechanisms that allow AI agents to assess their actions retrospectively, drawing lessons from successes and failures. Inspired by human cognition, these systems employ methods such as performance analysis, error detection, and strategic adaptation. Central to this is the concept of “deep processing,” wherein the deliberate examination of processes and outcomes enhances understanding and memory retention (Kosslyn, 2017). For AI agents, this means leveraging past experiences to refine their strategies, leading to improved task execution.

Reflection systems in AI often use frameworks such as reinforcement learning and causal reasoning to model the effects of decisions and/or actions in achieving desired outcomes. Recent advancements in agentic design patterns emphasize the value of structured reflection to optimize task planning, execution, and adaptability (DeepLearning.ai, 2024). By implementing self-assessment loops, AI agents can iteratively improve their responses to complex challenges, demonstrating enhanced capabilities over time.

### 4.1.1 Mechanisms of Reflection Systems

Reflection systems are underpinned by several key mechanisms that define their functionality and potential. These mechanisms include interactive feedback loops, verbal reinforcement, and external verification and self-correction. Each mechanism provides a distinct but complementary pathway for enabling AI agents to reflect, learn, and adapt. These mechanisms build on both internal evaluation processes and external tools, creating a hybrid structure to address diverse challenges such as error detection, adaptability, and knowledge gaps. Together, they enable AI agents to iteratively analyze their outputs, refine their processes, and adjust their strategies with increasing autonomy and accuracy.

#### 4.1.1.1 Interactive Feedback

Modern advancements in reflection systems focus on integrating external feedback mechanisms, such as the CRITIC framework (Gou et al., 2024). CRITIC represents a cutting-edge approach to reflection systems by empowering AI agents to self-correct through tool-interactive critiquing. This framework enables LLMs to validate and refine their outputs by engaging with external tools, including search engines, code interpreters, and calculators. By mimicking human behaviors of cross-referencing and interactive improvement, CRITIC enhances the ability of AI systems to identify and address errors in real-time.

A key feature of CRITIC is its “verify-then-correct” cycle, where external tools are used to evaluate the accuracy, coherence, and reliability of the AI’s initial outputs. Based on critiques generated through these interactions, the AI refines its responses in iterative loops. This method mirrors the metacognitive processes of self-assessment and strategic adaptation found in human cognition. For example, in tasks such as free-form question answering, CRITIC leverages search engines to verify factual accuracy. In mathematical reasoning, it employs code interpreters to validate the logic and correctness of programmatic solutions. Additionally, CRITIC has demonstrated its effectiveness in reducing toxicity in content generation by utilizing external evaluation tools to identify and mitigate harmful language.

The integration of CRITIC into reflection systems marks a significant evolution in AI design by emphasizing the importance of external feedback as a complement to internal processing. Traditional reflection systems rely primarily on self-generated assessments, which can be limited by the model’s inherent knowledge gaps or biases. CRITIC addresses this limitation by incorporating external sources of truth, creating a hybrid system that combines the strengths of internal metacognitive processes with the objectivity of external validation. This hybrid approach has proven to yield substantial improvements in both accuracy and adaptability across a variety of tasks where precise and unbiased outputs are essential, such as generating financial models or scientific analyses. By operationalizing a “verify-then-correct” cycle with external tools, CRITIC mirrors classroom practices of peer review and fact-checking, helping students develop metacognitive skills of self-assessment and evidence-based reasoning.

#### 4.1.1.2 Verbal Reinforcement

Frameworks like Reflexion (Shinn et al., 2023) utilize verbal reinforcement to guide agents in self-improvement. Reflexion agents process feedback as episodic memory, enabling them to iteratively adjust their actions and strategies, achieving state-of-the-art performance in diverse domains such as decision-making, reasoning, and programming (Shinn et al., 2023). Reflexion's key innovation lies in its ability to transform binary or scalar feedback into detailed, actionable insights that are stored as long-term memory. This memory enables agents to recognize patterns of errors and adapt their approaches over multiple iterations, significantly enhancing performance across various tasks, including sequential decision-making and complex reasoning.

Beyond theoretical advancements, Reflexion has shown particular promise in multi-step logical reasoning tasks, where verbal reinforcement allows agents to identify and address gaps in their reasoning chains. This mechanism is also critical in areas where adaptability is required, such as debugging software or conducting multi-stage workflows involving dependency tracking. Storing and reusing feedback as episodic memory, allows Reflexion to simulate formative assessment cycles, enabling students to iteratively refine their problem-solving strategies and build resilience in multi-step reasoning tasks.

#### 4.1.1.3 External Verification and Self-Correction

SELF-REFINE (Madaan et al., 2023) introduces an iterative feedback loop that allows LLMs to improve their outputs without external supervision. SELF-REFINE uses the same model to generate, critique, and refine outputs, iterating until a satisfactory result is achieved. This process is grounded in the principles of human self-reflection, where individuals improve their work through cycles of evaluation and revision. SELF-REFINE excels in tasks such as code optimization, dialogue generation, and sentiment reversal by ensuring that feedback is specific and actionable, directly addressing deficiencies in the initial outputs. Unlike traditional approaches requiring extensive training or reinforcement learning, SELF-REFINE operates within a single model and uses few-shot prompts to guide the refinement process.

A unique aspect of SELF-REFINE is its ability to achieve significant performance gains with minimal computational resources. For instance, in code optimization tasks, SELF-REFINE not only improves efficiency but also enhances readability by iteratively refining algorithms based on self-provided feedback. Similarly, in dialogue generation, it transforms generic or incomplete responses into engaging and contextually appropriate conversations. This iterative capability is particularly effective in workflows where incremental improvement is vital, such as creating high-quality translations, polishing creative writing, or refining policy documents.

A new innovative addition to reflection systems is the integration of advanced reasoning frameworks such as DeepSeek (DeepSeek-R1), an open-source AI model designed for complex problem-solving. DeepSeek employs reinforcement learning and cognitive reasoning strategies, enabling it to iteratively refine its output and perform tasks with heightened precision.

These capabilities are especially relevant to reflection systems, where iterative feedback and self-correction are central to improving performance over time.

DeepSeek ability to incorporate external feedback and prior knowledge aligns with the goals of reflection systems in education. For instance, its structured reasoning framework can enhance the functionality of educational tools by analyzing student responses, identifying logical inconsistencies, and refining explanations or feedback to better support learning objectives. This mirrors the iterative refinement seen in frameworks like CRITIC and SELF-REFINE but expands on them by incorporating a broader range of reasoning and external validation mechanisms.

By leveraging advanced reasoning, DeepSeek could strengthen reflection systems' capacity to adapt to diverse educational contexts. Its open-source nature also lowers barriers to implementation, providing a flexible platform for integrating reflection capabilities into intelligent tutoring systems, automated grading tools, or collaborative learning environments. DeepSeek exemplifies how cutting-edge AI models can deepen the scope of reflection systems by incorporating sophisticated reasoning, thereby promoting accuracy and adaptability in educational applications.

The combination of CRITIC, Reflexion, SELF-REFINE, and DeepSeek showcases the evolution of reflection systems in AI. Together, these frameworks highlight the transformative potential of integrating internal and external feedback mechanisms, enabling AI agents to achieve higher levels of adaptability, accuracy, and reliability. These mechanisms collectively drive innovation across technical and creative domains, demonstrating the potential of reflection systems to redefine complex problem-solving and iterative improvement tasks. As reflection systems continue to advance, their applications in education are likely to expand, fostering more intelligent and human-like AI systems.

### 4.1.2 Reflection Systems in Education

In the field of education, reflection systems have emerged as transformative tools, offering new opportunities to enhance teaching, learning, and assessment. By leveraging iterative feedback, verbal reinforcement, and external verification & self-correction mechanisms, these systems enable AI-driven tools to provide dynamic, personalized, and effective educational experiences.

Reflection systems are increasingly employed in intelligent tutoring systems (ITS), where they enable real-time adaptation to student needs. These systems monitor student interactions, detect areas of struggle, and refine instructional strategies accordingly. For example, an ITS powered by frameworks like SELF-REFINE analyzes student errors in mathematics and iteratively adjusts its explanations, tailoring them to the learner’s understanding. This approach not only deepens comprehension but also fosters metacognitive practices by prompting students to reflect on their problem-solving strategies (Rouzegar & Makrehchi, 2024; Madaan et al., 2023).

Another promising application is in feedback generation and refinement. Reflection systems such as SELF-REFINE iteratively enhance feedback quality, making it more actionable and personalized. For example, in automated essay scoring, these systems provide detailed critiques on grammar, structure, and argumentation, enabling students to refine their writing over multiple revisions. By aligning with principles of deliberate practice, where iterative feedback drives skill acquisition, these systems contribute to long-term learning and improvement (Madaan et al., 2023).

Reflection systems also play a vital role in collaborative learning environments, where they enhance group dynamics and ensure equitable participation. AI agents embedded in group projects analyze patterns of engagement, identify disparities in contributions, and recommend interventions to balance collaboration. For instance, tools based on Reflexion can monitor the flow of a group discussion, highlight under-participating members, and suggest prompts to ensure their active involvement (Shinn et al., 2023). These capabilities align with collaborative pedagogical practices, fostering accountability and collective success.

In addition, reflection systems are instrumental in promoting metacognitive skills among learners. By tracking attention, engagement, and learning behaviors, these systems encourage students to reflect on their strategies and optimize their performance. For example, platforms integrating CRITIC can analyze a student's engagement with learning materials and suggest adjustments to improve focus and retention. This reflective feedback helps learners develop self-regulation and deeper awareness of their learning processes (Cukurova, 2024; Gou et al., 2024).

In automated grading, reflection systems are utilized to ensure fairness and accuracy in assessments. By cross-referencing student responses with validated sources, these systems refine grading algorithms to minimize bias and errors. For instance, frameworks like CRITIC incorporate external tools such as search engines and knowledge graphs to validate grading decisions, ensuring consistency and objectivity. This iterative refinement process addresses common challenges in large-scale assessments, where uniformity and reliability are critical (Gou et al., 2024; Yesilyurt, 2023).

Educational simulations also benefit from reflection systems, which enable students to practice real-world skills in immersive, AI-driven environments. In tools like PitchQuest, a multi-agent educational simulation, reflection mechanisms allow students to refine their performance through iterative feedback from AI-generated mentors and evaluators. These systems provide tailored guidance, fostering experiential learning and skill development in simulated settings (Mollick et al., 2024).

Similarly, the EduAgent framework leverages generative AI to simulate student behaviors in educational environments, enabling both instructors and AI systems to refine their instructional methods. This system models diverse student personas, providing dynamic interactions that help educators test and optimize learning strategies. By incorporating reflection mechanisms, EduAgent can iteratively adapt its simulations based on real-time performance data, ensuring that interactions remain realistic and pedagogically effective. Such advancements highlight the role of generative AI in creating more adaptable and reflective learning simulations (XU et al., 2024).

DeepSeek further enriches the potential of reflection systems in education by introducing advanced reasoning capabilities tailored to complex educational settings. DeepSeek enables systems to analyze nuanced student behaviors, refine instructional strategies, and provide personalized feedback. For example, DeepSeek can identify logical inconsistencies in a student’s problem-solving approach and refine its explanations iteratively to enhance conceptual understanding. Its reinforcement learning framework supports dynamic adaptability, making it an invaluable tool for addressing individual differences in student learning styles and progress (Romero, 2024, DeepSeek, 2024).

By integrating DeepSeek reasoning capabilities, reflection systems can advance the development of educational simulations and ITS platforms, where real-time evaluation and adjustment are critical. This includes applications in automated essay scoring, where the system assesses nuances such as argument coherence and writing style. Moreover, DeepSeek's ability to adaptively model complex behaviors enables it to simulate realistic student personas, enriching teacher training and curriculum testing environments.

Reflection systems in education are a powerful means of personalizing instruction, fostering metacognition, and improving collaborative learning. As these systems continue to evolve, they will play a central role in creating adaptive, intelligent learning environments that align with the diverse needs of students and educators.

Table 4.1. Reflection Systems in Education.

| **Source** | **Domain** | **Summary** |
|---|---|---|
| Madaan et al. (2023) | Feedback Refinement | Introduce SELF-REFINE, a framework that iteratively enhances feedback quality for tasks such as essay scoring, dialogue generation, and code optimization. |
| Shinn et al. (2023) | Collaborative Learning | Propose Reflexion, which uses verbal reinforcement to refine actions iteratively, improving group participation and active learning in educational projects. |
| Cukurova et al. (2024) | Metacognitive Practices | Highlight how reflective systems track engagement and learning behaviors, encouraging students to optimize their learning strategies for better performance. |
| Gou et al. (2024) | Automated Grading | Present CRITIC, a system that integrates external verification tools to enhance grading accuracy and fairness, particularly in large-scale assessments. |
| Xu et al. (2024) | Educational Simulations | Propose EduAgent, a framework using generative AI to simulate realistic student behaviors, enabling teacher training and curriculum optimization. |
| Mollick et al. (2024) | Educational Simulations | Propose PitchQuest, a simulation platform where reflection systems provide iterative feedback to refine learners' performance in experiential learning settings. |
| DeepSeek (2024) | Intelligent Tutoring | Introduce DeepSeek, an open-source AI model using advanced reasoning and reinforcement learning to enhance ITS through interactive elements such as simulations, quizzes, and problem-solving exercises that engage students and enhance their understanding of complex concepts. |

### 4.1.3 Challenges and Future Directions

Despite their distinct advantages, multi-agent systems also present several challenges. One major challenge is computational complexity, which arises from the iterative nature of reflection systems. Frameworks like CRITIC, SELF-REFINE, and DeepSeek rely on repeated cycles of evaluation and refinement, making them resource-intensive and limiting their scalability in resource-constrained educational settings (Gou et al., 2024; Madaan et al., 2023). Addressing this challenge will require innovations in computational efficiency and system optimization.

Data privacy and security also remain critical concerns. Reflection systems analyze vast amounts of sensitive user data, including learning patterns and performance metrics, which must be safeguarded against misuse or unauthorized access. Adherence to regulations is essential to maintaining trust and ensuring the ethical use of AI in education (U.S. Department of Education, 2023). Thus, it is important to design systems that are both secure and transparent.

Another issue is the propagation of biases in feedback. Self-generated or externally sourced feedback loops may inadvertently reinforce errors or perpetuate biases present in training data or external verification tools. For example, tools like search engines or code evaluators integrated into reflection systems may introduce inaccuracies if they themselves contain biases or errors (Shinn et al., 2023; Bender et al., 2021). To mitigate this, future systems will need mechanisms to identify and address such biases proactively.

Reflection systems also face difficulties in generalizing across diverse educational contexts. While effective in specific domains, these systems often struggle to adapt to varied disciplines, cultural contexts, or instructional styles (Luckin et al., 2016). Ensuring flexibility and adaptability remains a critical area of research to enable broader deployment. Additionally, the lack of transparency in how reflection systems generate feedback raises ethical concerns. Students and educators need clear explanations of how these systems arrive at decisions to build trust and accountability in their use (Zhang et al., 2024a).

Despite these challenges, the future of reflection systems in education is promising. Integrating multimodal feedback sources, such as speech, handwriting, and gestures, could greatly enhance the depth and quality of feedback (Cukurova, 2024; Kosslyn, 2017). For instance, systems could analyze a combination of verbal explanations and written solutions to provide comprehensive feedback tailored to individual learners. Collaborative frameworks, where multiple AI agents engage in reflective dialogues or interact with human educators, are another promising direction. Such systems could leverage the complementary strengths of AI and human expertise to deliver more robust and reliable feedback (DeepLearning.ai, 2024).

Ethical alignment and pedagogical best practices will also play a pivotal role in shaping the future of reflection systems. Close collaboration between AI researchers and educators is essential to ensure that these systems are designed to be fair, inclusive, and supportive of diverse learning needs (Holmes et al., 2019). Moreover, balancing AI capabilities with human oversight will help prevent over-reliance on automated systems and ensure that critical thinking and metacognition remain integral to the learning process (U.S. Department of Education, 2023).

Finally, advancements in real-time adaptive feedback offer significant potential for education. For instance, during group activities, reflection systems could monitor team dynamics in real-time and provide actionable recommendations to improve collaboration (Mollick et al., 2024; Zhang et al., 2024b). These innovations, coupled with efforts to ensure transparency and address bias, will help reflection systems become more effective and equitable tools for education.

## 4.2 Planning Systems

To fully harness the power of LLMs in education, it is crucial to explore and implement agentic design patterns that enable AI agents to autonomously select options, plan, process, revise, solve, and execute complex tasks. By decomposing these complex tasks into smaller, more manageable subtasks—often involving interaction with external tools and resources (Hu & Clune, 2024; Hu et al., 2024)—educational systems can become both more efficient and more sustainable. This modular approach not only facilitates targeted interventions and personalized support for learners but also helps optimize resource use, thereby reducing computational overhead and contributing to long-term ecological viability.

Abuelsaad et al. (2024) showed for instance, that chain-of-thought (CoT) prompting significantly improved LLMs' multistep reasoning abilities and “explanations” (Turpin et al., 2024), breaking tasks “step by step” (Yao et al., 2024) and problems “into intermediate reasoning steps, guiding the model through logical processes to reach a final conclusion” (Zhang et al., 2024a, p. 6817), according to two approaches (Singh et al., 2024; Xu et al., 2023): a) decomposition and b) interleaved approach.

In decomposition, the LLM fully decomposes the primary task first into a set of constituent sub-tasks. Subsequently, the LLM generates independent sub-plans for each of these sub-tasks, outlining the specific actions required for their fulfillment before execution starts. Finally, these sub-tasks are executed sequentially, ensuring the completion of each sub-task in a predetermined order until the overarching task is achieved (Singh, 2024; Xu et al., 2023).

In the interleaved approach, decomposition, planning, and execution are intricately intertwined. Rather than conducting a complete initial decomposition of the overarching task, the LLM Agent begins with a partial decomposition, typically focusing on an initial sub-task. Concurrently, planning and execution pertaining to this sub-task are initiated. As each sub-task is addressed, the agent dynamically adapts by identifying and incorporating newly recognized sub-tasks into the planning and execution process. The iterative cycle of decomposition, planning, and execution fosters a degree of flexibility, enabling the agent to effectively respond to unforeseen environmental complexities or emergent changes (Singh, 2024; Xu et al., 2023). Two prominent approaches to operationalizing this process are ReACT and ReWOO, which also carry significant pedagogical implications.

ReACT emphasizes the iterative interplay between reasoning and acting. It involves the following steps: (1) Reasoning — the agent analyzes the current situation, identifies potential actions, and predicts their outcomes; (2) Acting — the agent executes the chosen action, interacting with the environment; (3) Observing — the agent perceives the consequences of its action, updating its understanding of the world; and (4) Reflecting — the agent evaluates the outcome, adjusts its plan if necessary, and repeats the cycle.

Pedagogically, the ReACT cycle mirrors effective inquiry-based learning: students hypothesize (reason), experiment (act), interpret evidence (observe), and engage in metacognition (reflect). Embedding this cycle into intelligent tutoring systems or classroom simulations can encourage students to practice iterative problem solving and self-regulated learning. For example, an agent designed with ReACT principles could guide students through scientific experiments, prompting them to predict outcomes, test hypotheses, and reflect on discrepancies. This aligns with constructivist pedagogy by making learning an active, adaptive process rather than a passive reception of information.

ReWOO, by contrast, integrates planning, working, and orchestrating. The planner generates a comprehensive plan that considers multiple constraints; the worker carries out interactions, tool use, and data gathering while adapting to new information; and the solver consolidates the plan and generates a final solution.

The pedagogical implication here is twofold. First, ReWOO operationalizes the principle of interleaving from cognitive load theory, where tasks are mixed rather than practiced in isolation. Just as interleaved practice challenges students to differentiate between problem types and strengthen long-term retention, ReWOO challenges the agent to navigate dynamic environments by flexibly mixing planning, execution, and adaptation. Second, ReWOO resonates with collaborative and project-based learning. The planner–worker–solver division can be mapped to roles in group, illustrating to students the value of distributed cognition and role differentiation in problem solving.
Both frameworks therefore highlight that pedagogical design can move beyond static content delivery toward dynamic cycles of reasoning, action, reflection, and adaptation. ReACT emphasizes micro-cycles of inquiry and reflection that build students' metacognitive awareness, while ReWOO emphasizes the benefits of interleaved and role-differentiated learning experiences that cultivate adaptability and collaborative thinking. In essence, both approaches embody the principle of breaking rigid instructional structures in favor of flexible, adaptive learning, thereby strengthening problem-solving, transfer, and resilience in complex educational contexts.

Planning systems are, therefore, a critical component of AI agents, enabling them to (1) Define clear objectives/goals/outcomes for the agent's actions; (2) Generate Plans and create a sequence of actions to achieve the goals; and (3) Execute the planned actions, "accomplish a task end-to-end" (Masterman et al., 2024, p.1), monitoring progress and adapting as needed.

In terms of planning, agentic planning systems can be categorized into the following five categories, as shown in Table 4.2.

Table 4.2. Agentic Planning Systems

| | |
|---|---|
| Reactive Planners | Reactive Planners lack explicit planning capabilities. They operate based on simple stimulus-response rules or pre-defined behaviors. They react to immediate changes in their environment without considering long-term goals or consequences. |
| Simple Planners | Simple Planners can perform basic planning by breaking down simple tasks into a sequence of actions. They may use simple search algorithms or rule-based systems to generate plans. However, they lack the ability to handle complex or dynamic environments. |
| Goal-Oriented Planners | Goal-Oriented Planners can define and pursue specific goals. They can generate and execute plans to achieve these goals, considering constraints and available resources. They may also be able to adapt their plans based on feedback or unexpected events. |
| Learning Planners | Learning Planners can learn from past experiences and improve their planning capabilities over time. They can adapt their planning strategies based on successes and failures, leading to more efficient and effective plans. (e.g. WebAgent, Gur et al., 2023) |
| Self-Improving Planners | Self-Improving Planners can not only learn from past experiences but also actively seek out new information and knowledge to improve their planning capabilities. They can continuously refine their planning algorithms and strategies, leading to increasingly sophisticated and autonomous behavior. |

The planning process is shown in Figure 4.1 and described step by step below:

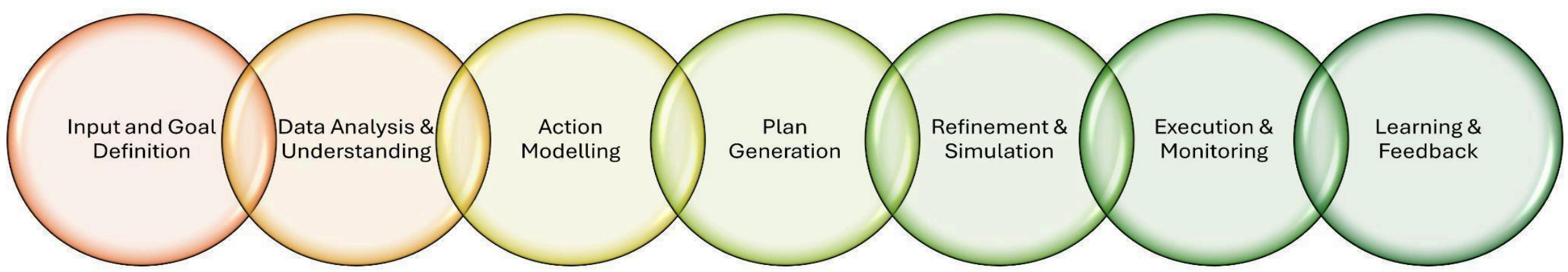

**Fig. 4.1** The sequence of the agentic planning process

1. Input and Goal Definition:
    - The agent clearly defines its desired outcome or objective.
    - This goal provides direction for the planning process.
2. Data Analysis & Understanding:
    - The input data is processed using machine learning to identify patterns and gather insights.
3. Action Modeling:
    - The agent identifies potential actions that can be taken to move toward the goal
4. Plan Generation:
    - The agent uses a planning algorithm (e.g., search algorithms like depth-first search, breadth-first search, or A* - heuristic search) to explore possible sequences of actions, considering constraints and possible outcomes.
5. Refinement & Simulation:
    - The agent may refine the plan based on feedback or new information, simulating outcomes to ensure efficiency and effectiveness.
6. Execution & Monitoring:
    - The agent executes the chosen plan, carries out the sequence of actions, and monitors its performance.
7. Learning & Feedback:
    - The agent learns from the outcomes and feedback (loop), improving its planning capabilities for future tasks.

The planning process, which involves the formulation of a sequence of actions to achieve specific goals, holds significant potential for transforming educational practices and outcomes. By leveraging the structured, adaptive, and goal-oriented nature of AI planning, educators can create personalized learning experiences, design curricula, optimize resource allocation, and enhance decision-making processes in educational settings. AI planning systems, which are designed to solve complex problems by breaking them down into manageable steps, can be applied to curriculum design, individualized learning pathways, and administrative tasks, thereby addressing some of the most pressing challenges in education today.

One of the most promising applications of AI planning in education is the development of personalized learning experiences (Kaswan et al., 2024; Sajja et al., 2024). Current educational models often adopt a one-size-fits-all approach, which fails to account for the unique learning preferences and paces of individual students. AI planning agents can address this limitation by analyzing vast amounts of data on student performance, engagement (in class or online), and behavior to create customized learning plans. To manage this data complexity effectively, agents can employ advanced mathematical methods in feature selection (Kamalov et al., 2025) to isolate the most predictive indicators of student success and assess a student's strengths and weaknesses in a particular subject, identify gaps in knowledge, and generate a tailored sequence of learning activities to address those gaps (Kikalishvili, 2024). This adaptive learning process ensures that students receive targeted support and challenges, thereby maximizing their learning potential (Kabudi et al., 2021). Moreover, AI planning can dynamically adjust these learning pathways in real-time based on ongoing assessments, ensuring the educational experience remains relevant and effective as students progress (Hyper-personalization). Research by Copyleaks (2024) indicated that 87% of educators and 78% of students believed that AI could revolutionize education through personalized learning experiences.

AI planning agents can further enhance personalization by generating detailed, constantly updated journey maps and reports for individual students or cohorts. These journey maps analyze and visualize the digital and physical multi-touchpoint experiences of students from enrollment to graduation, using techniques such as knowledge tracing (Shehata et al., 2023) and predictive modeling based on academic results, GPA, interests, selected study programs, study plans, and academic advisors' reports. Such comprehensive views of student journeys enable institutions to identify critical interaction points, align academic programs and support services with evolving student needs, facilitate collaboration across institutional departments, and support data-driven decision-making for recruitment, engagement, retention, and alumni relations.

In addition to personalization, AI planning can significantly enhance curriculum design by optimizing the sequencing and delivery of educational content. Curriculum development requires careful consideration of learning outcomes, prerequisite knowledge, constructive alignment, and logical progression. AI systems can analyze these factors and design coherent, efficient curricula. For instance, an AI planning agent can determine the optimal sequence for teaching topics, ensuring that students build foundational knowledge before tackling advanced material. This structured approach not only improves learning outcomes but also reduces cognitive load. Moreover, AI agents can continuously adapt curricula by automatically updating program maps when new courses are added, considering student-specific prerequisites and interests. They can also suggest electives from other programs aligned with career goals or research interests and align assessment tasks closely with specific learning outcomes, enabling comprehensive evaluation and performance predictions.

AI planning capabilities extend further into educational decision-making, providing insights and recommendations to support evidence-based practices. Educational decision-making often involves complex trade-offs and uncertainties, such as determining effective teaching strategies or assessing policy impacts. AI systems can analyze historical and real-time data to generate predictive models and scenario analyses (Hao et al., 2024). For example, an AI agent can simulate potential outcomes of various interventions or teaching methodologies, thereby recommending the most effective actions. Personalized recommendation systems that continuously learn from user interactions can help plan comprehensive teaching and learning activities that align with learning outcomes, student engagement, and assessment strategies (Yusuf et al., 2025).

Moreover, AI planning can foster collaborative learning environments, essential for developing critical thinking, communication, and teamwork skills. AI agents facilitate collaboration by organizing group activities, assigning roles based on individual student abilities, and monitoring progress. Systems like Bland AI provide real-time feedback and multilingual guidance, helping groups overcome challenges and remain goal-focused.

AI planning is also instrumental in optimizing resource allocation within educational institutions, addressing challenges related to limited resources like time, funding, and personnel. For instance, AI agents can analyze student enrollment data, faculty availability, and classroom capacity to create optimal timetables, allocate budgets effectively, prioritize projects, and streamline administrative processes. By automating these tasks, educators and administrators can concentrate on strategic initiatives and direct student engagement. Additionally, planning agents can select and organize educational resources tailored to individual student profiles, adjusting difficulty levels in real-time and visualizing information through mind maps. Such visual representations help students connect new information to their existing knowledge, thereby reducing cognitive load and enhancing learning (Hultberg & Calonge, 2017).

Finally, the integration of AI planning into education also has the potential to address issues of equity and access (Kamalov et al., 2023), which remain significant barriers to achieving universal quality education. AI systems can analyze data on student demographics, socioeconomic status, and geographic location to identify disparities in educational opportunities and outcomes. Based on this analysis, AI planning agents can design targeted interventions to support underserved populations, such as providing additional resources to schools in low-income areas or offering personalized tutoring to students who are at risk of falling behind. Furthermore, AI planning can facilitate the delivery of education in remote or resource-constrained settings (Kouam & Muchowe, 2025) through the use of digital platforms and adaptive technologies (Shah & Calonge, 2023). By leveraging AI to bridge gaps in access and quality (Aderibigbe et al., 2023), educators can work toward creating a more inclusive and equitable educational system.

Despite its numerous advantages, the implementation of AI planning in education is not without challenges. One of the primary concerns is the ethical use of data, as AI systems rely on vast amounts of personal information to function effectively. Ensuring the privacy and security of student data is paramount, and institutions must establish robust policies and safeguards to protect sensitive information. Additionally, there is a risk of algorithmic bias (Chan, 2023), where AI systems may inadvertently perpetuate existing inequalities or reinforce stereotypes. Addressing these risks requires robust data processing tools, such as analytical formulations for handling imbalanced datasets (Kamalov, Choutri, & Atiya, 2025), to ensure that minority groups are adequately represented in the training and inference phases. To mitigate/minimize this risk, it is essential to develop AI planning models that are transparent, accountable (Habbal et al., 2024), and regularly audited for fairness.

In conclusion, the planning process in AI agents offers a powerful framework for addressing some of the most pressing challenges in education, from personalizing learning experiences to optimizing resource allocation and enhancing decision-making. By leveraging the structured, adaptive, and goal-oriented nature of AI planning, educators can create more effective, efficient, and equitable educational systems. However, realizing the full potential of AI planning in education requires careful consideration of ethical, technical, and practical issues, as well as a commitment to fostering collaboration and innovation among all stakeholders.

## 4.3 Tool-use systems

In the domain of AI agents, the concept of "tools-use" refers to any functions or capabilities that the model can utilize to enhance its performance (Dwivedi et al., 2021). These tools enable agents to interact with external data sources, thereby facilitating the flow of information by retrieving or sending data and information to these sources. For instance, an AI agent designed as an educational tutor for high school mathematics. This persona is equipped with a deep understanding of mathematical concepts and the specific tasks it must accomplish, along with a suite of tools that facilitate various functions. These tools may include a graphing calculator for visualizing complex equations, a database of educational resources for providing supplementary materials, and a communication interface for sending personalized feedback to students via email or a messaging platform. The agent can analyze a student’s performance on practice problems, identify areas of difficulty, and then use its tools to generate tailored practice exercises or recommend videos and articles that explain challenging concepts.

A notable benefit of adopting an agent-based approach, as opposed to relying solely on prompting-based language models, lies in the agents' ability to address complex problems through the coordinated application of multiple tools (Shinn et al., 2023). Such tools empower agents to engage with external data sources, interact with existing APIs, and execute a variety of functions that significantly enhance their problem-solving capabilities. Tasks that necessitate extensive tool utilization are often intertwined with those that require intricate reasoning. Both single-agent and multi-agent architectures can effectively tackle challenging tasks by integrating reasoning and tool-calling mechanisms.

Many methodologies involve iterative processes of reasoning, memory utilization, and reflection to navigate and resolve problems with efficacy (Shinn et al., 2023; Shin et al., 2024). This often involves decomposing larger issues into smaller, more manageable subproblems that can be addressed sequentially using the appropriate tools, as explained above. Prior research (i.e., Dwivedi et al., 2021; Gao, 2024) indicates that while breaking down complex problems into smaller components can facilitate effective solutions, single-agent patterns frequently encounter challenges when faced with the lengthy sequences of actions required to complete these tasks.

### 4.3.1.Components and Mechanisms of Tool-Use Systems

The tool-use components and mechanisms in AI agents are fundamental for enabling dynamic interactions with external data sources and functionalities. This mechanism allows agents to execute complex tasks by leveraging various tools, thereby enhancing their capabilities to provide accurate and contextually relevant outputs. The following table summarizes the basic components and Processes of Tool-use in AI agents.

Table 4.3. Components and Mechanisms of Tool-Use Systems.

| **Components** | **Processes described** |
|---|---|
| Prompting and Tool Invocation | The agent initiates tool use through specially formatted requests.<br><br>Upon recognizing strings, a post-processing component triggers the appropriate tool, executes the function, and returns the results for further processing. |
| Iterative Reasoning | The tool use mechanism often involves iterative reasoning, where the agent reflects on previous outputs and adjusts its approach based on the information retrieved from external sources. |
| Content Management | Context management is crucial when multiple tools are available. The agent must determine the most relevant tools for the current task, often employing heuristics to prioritize tool selection. |
| Multi-Agent Collaboration-ready tool | In complex scenarios, multiple agents may collaborate, each employing specialized tools to address subproblems. This enhances efficiency and allows for parallel processing of tasks. |

The tool use mechanism in AI agents encompasses several key components and processes that enhance their functionality (see Table 4.3). Prompting and Tool Invocation begins the process, where the agent initiates tool use through specially formatted requests. Upon recognizing these strings, a post-processing component activates the appropriate tool, executing the designated function and returning results for further analysis. This leads to Iterative Reasoning, where the agent reflects on previous outputs, adjusting its approach based on new insights gained from external sources. Effective Content Management is essential as well, especially when multiple tools are available; the agent must discern the most relevant tools for the current task, often utilizing heuristics to prioritize tool selection. Finally, in more complex scenarios, multi-agent collaboration comes into play, allowing multiple agents to work together, each leveraging specialized tools to tackle subproblems. This collaborative approach not only enhances efficiency but also facilitates parallel processing of tasks, resulting in a more robust and adaptive system.

Indeed, "tools" are specific functions or capabilities that the AI agent can invoke. Each tool corresponds to a goal followed by an action, such as retrieving data, performing calculations, or interacting with third-party applications.

Table 4.4. Tool-use System (Operational Framework) Example

| **Steps** | **Description of Goals and Actions** | **Examples of Tools and Models used** |
|---|---|---|
| 1. Tool Initiation | The session begins with the agent receiving a task or query that necessitates tool use | For example, user input prompts - "*Who is the program leader XXX of University of XXX, analyses his/ her publications and send an "Happy New Year" email to him/her and wish the person good luck in his/her career*".<br><br>- "*Visualize the class composition with simple statistics and figures*" |
| 2. Tool Selection | The agent evaluates available tools based on the task's context and formulates a plan for invoking the necessary functions | · Web-search (e.g. Google Search API for retrieving current information)<br>· SMTP Email Sender for sending automated emails<br>· OpenCV Library for performing image recognition<br>· SQL Database Connector to retrieve user-specific information. |
| 3. Execution of Tool Functions | The agent constructs tool invocation strings and executes the associated functions | · Sending requests to APIs (e.g., Twitter API for social media interactions).<br>· Code execution tool: executing Python code |
| 4. Data Processing and Reflection | Once the tools return results, the agent processes the incoming data, incorporating it into its understanding of the task. The agent may then reflect on this information, iterating its reasoning to refine its output. | · Pandas: A library for data manipulation and analysis in Python. An AI agent can use Pandas to clean, transform, and analyze datasets.<br>· NumPy: A library for numerical computing in Python. An agent can utilize NumPy for performing mathematical operations on large arrays and matrices.<br>· Apache Spark: A unified analytics engine for large-scale data processing. An AI agent can use Spark to handle big data processing tasks efficiently. |
| 5. Output Generation | The agent synthesizes the information gathered through tool use and generates a coherent response or action. This output is informed by both the initial task and the insights gained from the tools | · Matplotlib for visualizing data<br>· TensorFlow: An AI agent can use TensorFlow to retrain models based on new data or feedback, reflecting on previous performance.<br>· Scikit-learn: An agent can evaluate model performance and adjust parameters based on reflection from past predictions.<br>· Jupyter Notebooks: An interactive computing environment where agents can document their processing steps, visualize data, and reflect on the outcomes of their analyses. |

The operational framework of AI agents (see Table 4.4) involves a series of structured steps designed to effectively address user tasks through tool utilization. Tool Initiation marks the beginning of this process, as the agent receives a task or query that necessitates tool use. For example, a user might prompt the agent with, "Who is the program leader of XXX in the University of YYY, analyze his/her publications, and send a Happy New Year email to her, wishing him/her good luck in her career," or "Visualize the class composition with simple statistics and figures." Following this, in the Tool Selection phase, the agent evaluates the context of the task and formulates a plan for invoking the necessary functions, considering tools such as the Google Search API for retrieving current information, SMTP Email Sender for sending automated emails, OpenCV Library for performing image recognition, and SQL Database Connector for accessing user-specific data. In the Execution of Tool Functions step, the agent constructs tool invocation strings and executes the associated functions, which may include sending requests to APIs like the Twitter API for social media interactions or executing Python code for data analysis. After the tools return results, the agent enters the Data Processing and Reflection phase, where it processes the incoming data to deepen its understanding of the task. This can involve using libraries like Pandas for data manipulation (Wang et al., 2024), NumPy for numerical computations, or Apache Spark for large-scale data processing. Finally, in the Output Generation step, the agent synthesizes the information gathered through tool use and generates a coherent response or action, informed by both the initial task and insights gained. This output may leverage tools like Matplotlib for data visualization (Wang et al., 2024), TensorFlow for model retraining based on new data, Scikit-learn for evaluating model performance, and Jupyter Notebooks for documenting processing steps and reflecting on outcomes. Through these steps, AI agents effectively navigate complex tasks, utilizing a variety of tools and models to deliver accurate and contextually relevant results.

## 4.3.2. Type of Tool-Use Systems

Aforementioned, within the tool-use systems paradigm, tools are essential for enhancing the capabilities of AI agents, allowing them to perform complex tasks effectively. These tools can be categorized into several groups based on their functionalities and applications. In Table 4.5 below, we outline key types of tool-use systems, providing descriptions and examples for each type.

Table 4.5. Type of Tool-use System

| **Type of Tools** | | **Descriptions** | **Examples** |
|---|---|---|---|
| **Data and Information Processing Tools** | Data Processing and Analysis | Focuses on the collection, manipulation, and analysis of data to extract meaningful insights. They enable AI agents to handle large datasets and perform data-driven decision-making. | Pandas: data manipulation, Apache Spark : large-scale data processing, Google search engine APIs for web enquiries |
| | Natural Language Processing (NLP) | Empowers AI agents to understand, interpret, and generate human language, facilitating communication between the agent and users. They enhance the ability to process textual information effectively | SpaCy: text processing, NLTK: linguistic data analysis |
| | Computer Vision | Enables AI agents to interpret and analyze visual information from the world, allowing for tasks such as image recognition and video analysis, which are crucial in simulations and interactive environments. | OpenCV: image processing, TensorFlow: deep learning applications in vision<br><br>Yolo: object detection algorithm |
| **Planning, Modeling and Decision Making Tools** | Decision Making and Planning Tools | Assists AI agents in making informed decisions and planning complex tasks by analyzing various factors and potential outcomes. | IBM Decision Optimization for planning and scheduling, Microsoft Power BI for business intelligence. |
| | Simulation and Modeling | Allows AI agents to create and manipulate models of real-world scenarios for training, assessment, and simulations. They foster experiential learning opportunities. | AnyLogic: simulation modeling, MATLAB: mathematical modeling |
| **Interaction and Communication Tools** | Interaction and Communication | Facilitates interactions between AI agents and users or between multiple agents, enhancing collaboration and feedback mechanisms. | Chatbot frameworks such as Rasa for conversational interfaces, Zoom API: virtual meetings. |
| **Governance and Evaluation Tools** | Monitoring and Evaluation | Helps assess performance and engagement metrics, providing insights into user interactions and outcomes. | Google Analytics for tracking user engagement, formative assessment tools for real-time feedback. |
| | Ethics and Fairness | Focuses on ensuring that AI systems operate within ethical guidelines, promoting fairness and accountability in decision-making processes. | Fairness Indicators for evaluating model fairness, AI ethics frameworks for compliance with ethical standards. |
| **Development and Deployment Tools** | Development and Collaboration | Supports the collaborative development of AI applications, allowing teams to work together effectively while building and deploying AI systems. | GitHub for version control and collaboration, Jupyter Notebooks for interactive coding and sharing of AI models. |
| | Deployment and Scaling | Assists in the deployment of AI applications and their scaling to accommodate larger user bases or more complex tasks, ensuring efficiency in operation. | Docker for containerization, Kubernetes for orchestration of containerized applications. |

By utilizing a combination of these types of tools, AI agents can effectively process data, interact with users, and create meaningful learning experiences. As the integration of AI in education continues to evolve, understanding these tool-use systems in the educational settings will be critical for optimizing their application and maximizing the benefits for learners and educators alike.

### 4.3.3. Tool - Use systems in Education

Tool-use systems in education leverage the capabilities of AI agents to enhance learning experiences through dynamic interactions with various educational resources. These systems facilitate personalized and effective learning environments by coordinating multiple tools tailored to specific educational tasks (Chin et al., 2010; Ouyang & Jiao, 2021). For example, AI agents can utilize adaptive learning platforms and educational APIs to provide real-time feedback and support individualized learning paths, aligning with the AI-supported, learner-as-collaborator paradigm (Ouyang & Jiao, 2021).

One significant advantage of tool-use systems is their ability to actively engage learners. By integrating tools for document annotation, quiz generation, and performance analytics, AI agents assist educators in tracking student progress and adapting instructional strategies accordingly. This collaborative approach fosters a co-creation of knowledge between AI systems and students, thus enhancing the learning process (Holmes et al., 2019; Luckin & Cukurova, 2019). Additionally, these systems provide substantial benefits to teachers. AI agents can automate administrative tasks such as grading and attendance tracking, allowing teachers to focus more on instructional activities and student engagement. By analyzing student performance data, AI can also offer insights into class trends and individual student needs, enabling teachers to tailor their instruction more effectively. This data-driven approach helps educators identify which students may require additional support, ultimately improving learning outcomes (Luckin et al., 2016). Moreover, the iterative reasoning capabilities inherent in tool-use systems allow AI agents to reflect on previous interactions and adjust their strategies based on the evolving needs of learners. This reflective practice is essential in educational contexts, as understanding a student's unique learning trajectory can lead to more effective interventions (Zawacki-Richter et al., 2019). For instance, utilizing data analytics tools enables AI agents to identify patterns in student performance data, recommending personalized resources that cater to individual strengths and weaknesses.

Table 4.6 below highlights the relevant literature of tool-use systems in education.

4.6 Tool-use systems in Education

| **Author(s)** | **Summary** | **Relevant Tool-Use type** |
|---|---|---|
| Sellar & Gulson (2020) | Analyzed AI's impact on education policy and governance. The tool used brought in policy implications of AI in education. | Ethics and Fairness tools |
| Winters et al. (2020) | Explored digital structural violence in education and strategies to combat it. The tool assisted in implementation of strategies addressing inequities in educational technology. | Ethics and Fairness tools |
| Ouyang & Jiao (2021) | Identified three paradigms of AI in education: AI-directed (Intelligent Tutoring Systems), AI-supported (Dialogue-based Tutoring Systems), and AI-empowered. | Communication and Interaction tools |
| Barua (2024) | Explored LLMs' potential in creating autonomous agents with various capabilities across domains. LLMs used for tasks like customer service, healthcare diagnostics, and email responses | Simulation and Modeling tools |

### 4.3.4. Limitations and Challenges

Despite the advantages, the implementation of tool-use systems in education faces several limitations and challenges. One significant issue is the seamless integration of various educational tools and platforms. The literature emphasizes the necessity for robust interoperability standards to ensure smooth data flow between systems (Pinkwart, 2016). Without these standards, the effectiveness of tool-use systems can be significantly hampered, leading to fragmented learning experiences. More importantly, concerns about data privacy and the ethical implications of automated decision-making (Hultberg et al., 2024; Kamalov et al., 2023) in educational settings are critical challenges. As AI systems collect and analyze vast amounts of student data via multiple tools , ensuring the security and privacy of this information becomes paramount (Kamalov et al., 2023). The potential for bias in AI algorithms and tools also raises ethical questions about fairness and equity in educational outcomes (Ouyang & Jiao, 2021). Another challenge is the need for educators to adapt to these new tools and technologies by upskilling and reskilling (Santandreu Calonge et al., 2025). Many educators may lack the necessary training and support to effectively implement and integrate AI-driven tools into their teaching practices (Zawacki-Richter et al., 2019). This gap can lead to resistance to adopting AI tools, limiting their potential benefits in the classroom.

To tackle these challenges, educational institutions must develop comprehensive frameworks that guide the implementation of AI tool-use systems. This includes establishing clear data management guidelines, fostering collaboration between educators, students and AI developers (Santandreu Calonge et al., 2025), and ensuring alignment with pedagogical goals (Ouyang & Jiao, 2021). By prioritizing these aspects, educational stakeholders can harness the potential of tool-use systems to transform, ‘super-personalise’ learning experiences (Popenici, 2023), and promote learner agency. In conclusion, tool-use systems in education represent a transformative approach, enabling AI agents to enhance engagement, support personalized learning, and facilitate collaborative practices. As these systems continue to evolve, their effective implementation will be critical in shaping the future of education, making it more responsive to the diverse needs of learners (Ouyang & Jiao, 2021).

## 4.4 Multi Agent Systems

Recent advancements in natural language processing (NLP) have led to the emergence of highly capable agents that draw on LLMs for complex reasoning, tool usage, and real-time adaptation to novel observations (Xie et al., 2023; Wang et al., 2023b). As the variety and complexity of tasks suited to LLMs continue to grow, an effective approach to amplifying agent performance is to employ multiple agents working in tandem. Beyond improving efficiency, such collaboration also offers pathways to developing more sustainable AI-driven systems. Previous research shows that multi-agent approaches can foster divergent thinking (Liang et al., 2023), enhance factual accuracy and reasoning (Du et al., 2024), and provide systematic validation (Wu et al., 2024).

A key motivation for using multi-agent systems is the capacity of chat-optimized LLMs to comprehend and respond to feedback. When multiple LLM-based agents communicate—either with one another or with human collaborators—they can exchange reasoning steps, observations, critiques, and validation through dialogue. In addition, LLMs have demonstrated the ability to decompose complex tasks into smaller, more manageable subtasks. Multi-agent dialogues can facilitate task partitioning and subsequent reintegration in an intuitive manner that allows a more efficient and organized problem-solving process.

In contrast, single-agent systems face several limitations. First, when a single agent is equipped with multiple tools, it may struggle to make optimal decisions about which tool to use for a specific task. Second, as task complexity escalates, an overwhelming amount of contextual information can exceed the capacity of a single agent to manage effectively. Third, many real-world scenarios require multiple areas of specialization, which a single agent may not fully encompass. By dividing tasks among specialized agents with diverse expertise, a multi-agent paradigm can mitigate these issues and maintain more manageable contexts. As a result, multi-agent systems offer a promising avenue for dealing with the increasingly complex and specialized tasks in real-world applications.

Multi-agent approach presents several practical advantages, including modularity, specialization, and control. While the multi-agent approach is similar to the “planning” paradigm, it provides greater control over individual components of the system. Increased control and modularity allow to reduce the amount of hallucinations which is a major challenge in LLM-based applications. A summary of the advantages of the multi-agent approach is presented in Table 4.7.

Table 4.7 Advantages of multi-agent systems.

| **Benefit** | **Description** |
|---|---|
| Modularity | Using separate agents simplifies the development and maintenance of agentic systems by allowing isolated updates or replacements without disrupting the entire system. |
| Specialization | Dedicated, expert agents can be assigned to specific domains, boosting overall performance through focused expertise. |
| Control | Clear and explicit mechanisms for agent interactions (e.g., conversations) enable better oversight rather than relying solely on function calls. |

### 4.4.1 Multi Agent Architectures

There exist several architectures for implementing the multi-agent approach. In the “supervisor” architecture, a lead LLM delegates tasks to the subagents in the system. In the “network” architecture, each agent can communicate with another agent in the system. Table 4.8 below outlines several common multi-agent designs, along with their defining characteristics and potential challenges.

Table 4.8. Multi-agent architectures

| **Architecture** | **Key Characteristics** | **Potential Challenges** |
|---|---|---|
| Single (Baseline) | - Utilizes a single LLM that can call multiple tools. - Represents the simplest setup. | - Subject to previously noted limitations such as overwhelmed context and insufficient specialization. |
| Network | - Composed of numerous agents, each with its own tools and the ability to call upon other agents. - Often employed by projects such as Swarm or Crew AI. - Lacks a designated “leader,” allowing any agent to invoke any other agent at any time. | - Less predictable behavior. - Potentially expensive due to numerous LLM calls. - Time-consuming and complex to manage in production. |
| Supervisor | - Utilizes a single “supervisor” agent that decides which sub-agent to call next. - Sub-agents focus on specific domains, enhancing specialization. - A simplified variant treats sub-agents as tools used by the main agent, where the main agent’s tool calls handle inter-agent communication. | - Over-reliance on the supervisor agent for optimal orchestration. - Potential bottleneck if the supervisor agent is overwhelmed. |
| Hierarchical | - Builds on the supervisor model by introducing multiple layers of supervisor agents. - Allows for large, domain-specific systems to be organized in tiers. | - Complexity increases with each additional supervisory layer. - May require extensive coordination across tiers. |
| Custom | - Tailored specifically to the requirements of a given domain. - Often combines elements from supervisor or hierarchical models with custom domain logic. | - Development can be resource-intensive due to bespoke design. - Requires ongoing maintenance for evolving domain needs. |

### 4.4.2 Multi Agent Systems in Education

Multi-agent systems have been increasingly employed in various educational contexts (Table 4.9). They are particularly beneficial in complex tasks that demand the coordination of multiple components. In adaptive learning environments, for instance, individual agents can address distinct facets of the learning process—such as identifying knowledge gaps, adjusting content difficulty, or providing personalized feedback—thereby accommodating diverse learning styles and paces.

The use of multi-agent systems in education predates the advent of LLMs (Apoki et al., 2022). Viswanathan et al. (2022) propose a multi-agent system consisting of several specialized agents: Jade Gateway Agent, Control Agent, Learning Style Detector Agent, Adaptive Course Organizer Agent, and User Interface Agent. Similarly, Ivanova et al. (2023) present a design based on Learner Agent, Content Agent, and Pedagogical Agent to construct a personalized e-learning system. Similarly, Axac et al. (2023) propose a multi-agent system that integrates computer vision to construct an online learning platform. One promising avenue for further development involves integrating LLMs into the existing multi-agent frameworks.

Educational simulations represent another active area of multi-agent design application. Developing simulations can be both resource-intensive and time-consuming; however, advancements in generative AI provide potential solutions to these challenges. Mollick et al. (2024) propose “PitchQuest,” a multi-agent AI system for educational simulations. The system offers personalized learning experiences by allowing students to practice skills in scenarios populated by AI-generated stakeholders. The proposed prototype features multiple agents, including Mentor Agent, Investor Agent, Evaluator Agent, Progress Agent, and Class Insights Agent. In medical education, Wei et al. (2024) propose “MEDCO,” a multi-agent system that simulates realistic training environments by incorporating roles such as patient, doctor, and radiologist. Their design promotes interactive, multidisciplinary learning and enhances students’ question-asking and collaboration skills in a virtual setting (Williams & Li, 2023). SimClass is a multi-agent classroom simulation framework that mimics traditional classroom interactions by integrating real user participation and enabling collaborative, role-based teaching dynamics (Zhang et al., 2024b). Recent research suggests that multi-agent systems can offer more personalized and effective learning experiences by adapting to individual student needs (Chen & Zhao, 2023).

General AI-based methods for automated grading have also attracted scholarly attention (Wongvorachan et al., 2022; Yang et al., 2025). While multi-agent systems for automated grading have been explored to a limited extent, Guo et al. (2024) describe a system with two AI agents: one generates feedback, and the other validates and refines it. This approach significantly reduces instances of over-praise and over-inference, thereby yielding more accurate and pedagogically sound feedback. Yesilyurt (2023) similarly investigates the potential of multi-agent assessment and feedback mechanisms within language learning contexts, further underscoring the promise of multi-agent systems in enhancing educational processes.

Table 4.9. Multi-agent Approaches in Education

| **Source** | **Domain** | **Summary** |
|---|---|---|
| Mollick et al. (2024) | Simulation | Propose PitchQuest, a venture capital pitching simulator. Enables personalized learning through AI-generated stakeholders and includes multiple agents (Mentor, Investor, Evaluator, Progress, Class Insights). |
| Wei et al. (2024) | Simulation | Introduce MEDCO, a multi-agent system that simulates real-world training environments by incorporating patient, doctor, and radiologist roles, fostering interactive, multidisciplinary learning in a virtual setting. |
| Zhang et al. (2024b) | Simulation | Propose SimClass, a multi-agent classroom simulation framework that mimics traditional classroom interactions by integrating real user participation and enabling collaborative, role-based teaching dynamics. |
| Jin et al. (2024) | Simulation | Propose TeachTune, a simulated student agent system to test and tune the performance of pedagogical conversational agents |
| Ma et al. (2024) | Simulation | Propose AI4Education framework for modeling virtual student agents used for teacher training. |
| Guo et al. (2024) | Automated Grading | Develop a multi-agent system featuring two AI agents—one generating feedback and the other validating/refining it—for automatically scoring student responses with the aim of reducing over-praise and over-inference errors. |
| Yesilyurt (2023) | Automated Grading | Explore multi-agent assessment and feedback mechanisms in language learning contexts, emphasizing how multi-agent systems can support more nuanced and effective student feedback. |

### 4.4.3 Limitations

Despite their distinct advantages, multi-agent systems also present several challenges. Compared to single-agent approaches, coordinating multiple agents typically increases token usage and processing time. The increased resource consumption amplifies computational costs and environmental impact. Therefore, an important research goal is minimizing the number of interactions required while maintaining agent performance. Another key issue is effectively chaining and orchestrating individual agents in the system, as improper coordination can diminish the overall system's performance. Finally, given the propensity for LLMs to hallucinate, relying on communicative agents to handle tasks may lead to incomplete or inaccurate outputs (Agnihotri & Chug, 2020).

Multi agent systems can be complex to develop and require careful coordination to ensure agents remain aligned with the intended instructional goals. Without proper design, an agent-based system may suffer from inconsistencies. A breakdown in communication among agents can lead to learner confusion or reduced engagement. Biases in agent decision-making can skew interactions, and complex simulations may burden the system. Since agent-based interactions vary, students may receive significantly different experiences, which can undermine the fairness and consistency of instruction.

## 4.5 Discussion

Our analysis of the four agentic paradigms—reflection, planning, tool use, and multi-agent collaboration—reveals a significant conceptual shift in the role of AI in education. While presented as distinct, these paradigms are deeply interconnected, collectively moving beyond the single-function, content-delivery focus of historic pedagogical agents (e.g., Johnson et al., 2000). The synthesis of these paradigms frames a transition from AI as a static repository of knowledge to an autonomous partner in the learning process, capable of orchestrating complex educational workflows. It has profound implications for educational practice, suggesting a move away from AI tools that simply provide answers towards agentic systems that manage processes, facilitate inquiry, and adapt dynamically. For educators, it signals a potential role change from content experts to designers and facilitators of AI-mediated learning environments. For learners, it promises more personalized and agentic experiences. However, this increased autonomy and complexity introduces critical challenges. As the paradigms converge, the resulting agentic systems become more powerful but also less interpretable. The nested processes of a multi-agent system that plans, uses tools, and reflects on its own output can become a "black box," raising significant concerns about trustworthiness, accountability, and pedagogical alignment that must be at the forefront of future research.

The technical complexity inherent in the agentic paradigms is not merely an engineering choice; it carries direct pedagogical weight. For instance, the iterative, self-correcting loop in reflection frameworks like CRITIC and SELF-REFINE does more than improve output accuracy—it operationalizes the principles of formative assessment and deliberate practice for the learner. The technical separation of planning and execution in systems like ReWOO is not just for efficiency; it allows for the design of learning pathways that can strategically interleave concepts, a practice known to enhance long-term retention and combat the 'illusion of mastery' that often arises from blocked practice. By failing to examine these pedagogical implications, we risk developing technically sophisticated tools that are pedagogically naive. The true transformation promised by agentic AI in education lies in its ability to embed these proven learning principles into the very architecture of the system, creating environments that don't just provide information but actively cultivate the cognitive and metacognitive skills necessary for lifelong learning.

The pedagogical benefits of CRITIC, Reflexion, ReWOO and other technologies are contingent on several practical conditions. Key requirements include reliable infrastructure and platform interoperability, teacher professional development to interpret and mediate agentic feedback, and transparent, explainable outputs to ensure feedback is usable and accountable. Empirical claims about long-term learning, transfer, and equity should be validated through iterative pilot studies combining quantitative outcomes with classroom observation before large-scale roll-out. Finally, deployment must meet baseline ethical and governance standards so that technical sophistication translates into safe, interpretable, and educationally meaningful practice.

# 5 Illustrative Application: Multi-agent scoring system

To provide a concrete example of how agentic paradigms, specifically multi-agent collaboration, can be applied in an educational context, we developed a proof-of-concept Multi-Agent Scoring System (MASS) for automated essay grading. Essay writing is a fundamental skill, yet manual grading is time-consuming and potentially subjective. Automated Grading Systems (AGS) aim to address these issues, but single-LLM approaches can suffer from inconsistency and biases like over-praise (Guo et al., 2024).

As a proof of concept for the use of AI agents in education, we propose an automatic essay grading system based on a multi-agent approach. The proposed system consists of two independent subagents that are supervised by the lead agent (Fig 5.1). The first subagent scores an essay based on the content. It checks how well the essay is organized, clear coherence and progression of ideas using appropriate examples, reasons, and other evidence to support its position. The second agent checks if the essay is free of errors in grammar, usage, typos, and mechanics. The lead agent combines the feedback from each subagent to produce the final score. By focusing each subagent on a specific facet of essay writing, we aim to obtain more accurate feedback. Furthermore, to increase grading consistency, the lead agent is conditioned to request multiple reports from the subagents in case of widely divergent feedback. In practice, this mechanism was triggered in approximately 14% of essays where initial sub-agent scores differed by more than two points on the 1–6 rubric scale. Divergent cases most often reflected essays with strong ideas but weak grammar, or essays with polished language but limited argumentation. The Supervisory Agent's synthesis of multiple sub-agent reports helped balance such discrepancies to produce more reliable final scores.

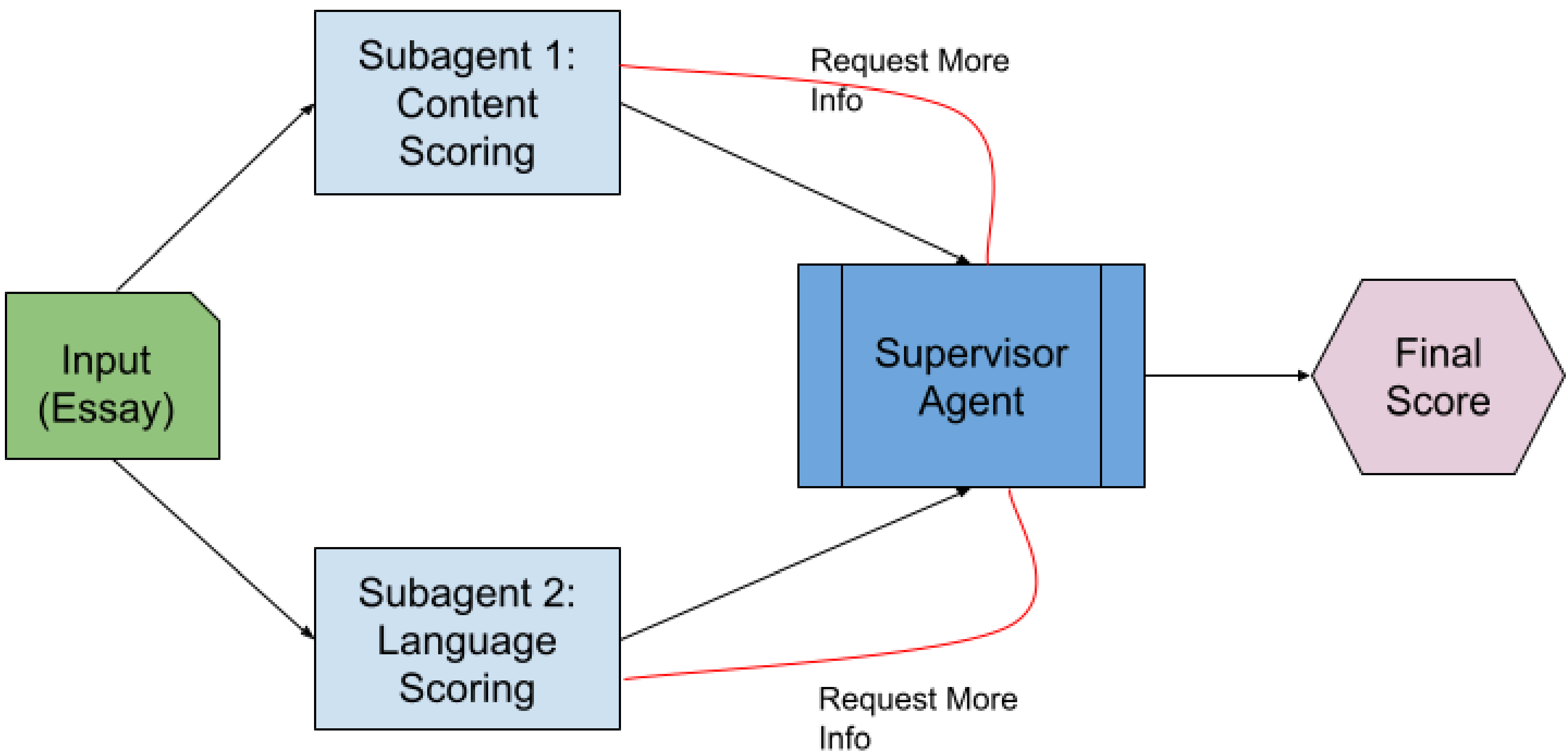

**Fig. 5.1** The proposed multi-agent assessment system (MASS) for automated essay grading.

The proposed multi-agent scoring system (MASS) is based on GPT-4o. We benchmark the MASS against several stand-alone LLMs: GPT-4o, DeepSeek 67b, and DeepSeek 1.3b. This comparative approach aligns with recent studies in other educational domains, such as entrepreneurship, which have demonstrated significant performance variability among leading AI chatbots depending on the specific application context (Kamalov et al., 2025). We use the ASAP 2.0 Dataset for our evaluation (Crossley et al., 2024). The dataset comprises about 17000 student-written argumentative essays. Each essay was scored on a scale of 1 to 6 based on a holistic rubric. The code and the technical details of the evaluation are available on GitHub (https://github.com/AzizovDilshod/Multi-agent-System-for-Essay-Assessment/tree/main).

Table 5.1 Comparison of MASS vs single LLMs in automatic essay grading.

| Model | MAE | Std Dev of Error |
|---|---|---|
| DeepSeek 1.3B | 1.696 | 1.767 |
| DeepSeek 67B | 0.735 | 1.010 |
| GPT-4o | 0.613 | 0.956 |
| Llama 3.3 70B | 0.614 | 0.783 |
| MASS | 0.561 | 0.830 |

The results of the evaluation are summarized in Table 5.1. The results show the advantages of the proposed MASS based on GPT-4o over stand-alone LLMs for automated essay grading. The MASS achieves the lowest mean absolute error (MAE) of 0.5612, indicating that its predictions align more closely with the ground truth scores than those of single LLMs. In contrast, the stand-alone GPT-4o model has a higher MAE of 0.6129, while the DeepSeek models perform worse, with DeepSeek 67b at 0.7345 and DeepSeek 1.3b at 1.6956. Moreover, the statistical tests show that the p-values for both the paired t-test and Wilcoxon signed-rank test are 0.0 across all comparisons. It indicates that the differences in MAE between MASS and each of the single-LLM models are highly statistically significant. We also note that Llama 3.3 70B produced a lower standard deviation of error compared to the MASS, indicating tighter clustering of its predictions.

The superior performance of MASS can be attributed to its multi-agent architecture, where separate subagents focus on different aspects of essay quality—content and language mechanics—before the lead agent synthesizes their outputs. The structured approach reduces errors stemming from over-praise and over-inference, which are common pitfalls in single-LLM grading. Additionally, the verification mechanism introduced in MASS can lead to more refined and balanced scoring.

# 6 Conclusion

In this paper, we explored how emerging AI agents can address the limitations of conventional LLMs in education. Our analysis was based on four key design paradigms—reflection, planning, tool use, and multi-agent collaboration. The survey of existing literature revealed the potential for the agentic workflows to offer greater adaptability, enhanced reasoning, and more consistent performance in educational settings. Furthermore, the proposed proof-of-concept multi-agent essay scoring system demonstrated the advantages of agentic workflows over stand-alone LLMs.

By incorporating real-time data, iterative feedback loops, and strategic task decomposition, AI agents overcome the static nature of traditional LLMs, making them powerful reasoning 'partners' or assistive tools. The new paradigm enables the delivery of more robust and trustworthy AI-driven solutions for educational needs in a variety of tasks. In particular, the multi-agent essay-scoring framework highlights how collaborative systems can improve reliability, consistency, and scalability of AI systems in education.

Despite our findings, integrating agentic workflows into diverse educational contexts holds several unresolved challenges. Achieving transparency in decision-making processes, ensuring fairness and accountability, and establishing ethical guidelines remain notable concerns. Designing systems that reliably transfer across heterogeneous learner populations and different pedagogical settings also warrants further investigation. Finally, the interplay between human instructors, learners, and AI agents must be systematically studied to ensure that agentic solutions support rather than supplant meaningful educational interactions.

## 6.1 Future Research

There are several avenues for future research to address the remaining challenges in deployment of AI agents in education. These include refining AI architectures to improve interpretability, exploring various strategies that better align with specific pedagogical goals, and developing policy frameworks to ensure fair and equitable deployment of AI agents in classroom settings. To address skepticism and resistance from stakeholders, including teachers, parents, and policymakers, research should focus on making AI agent reasoning transparent to educators and students (Explainable AI (XAI)).

Educational environments are dynamic and new technologies are constantly being introduced. Future research could focus on developing AI agents that are adaptive, context-aware, and can adapt to changing student needs, languages, culture, learning preferences, context (s) (disadvantaged, refugee, remote, etc.) to avoid exacerbating the digital divide. It is also important to study the effectiveness of low-cost or open-source AI solutions in under-resourced schools and communities. Additionally, more longitudinal research is needed to understand the long-term effects of AI on student learning outcomes, skills, teacher practices, and the overall educational ecosystem. It is crucial to investigate whether prolonged exposure to AI agents affects students' ability to be creative, engage in self-directed learning or collaborate with peers. Finally, closer cooperation between AI developers and educators is required to achieve better systems that take into account the needs of the students and instructors.

We recognize that environmental, infrastructural, and social sustainability are important factors in debates about large-scale deployment of AI systems. However, a rigorous treatment of sustainability requires dedicated empirical analysis and policy framing that extend beyond the conceptual scope of this review. Therefore, while we acknowledge sustainability as a relevant cross-cutting concern, we treat it as a priority area for follow-up research rather than as a core focus of this manuscript. Future work should evaluate the full lifecycle costs and equity implications of agentic systems, and develop benchmarks and governance instruments that make sustainability a measurable criterion for educational deployments.

## Declarations:

- Funding: No Funding.
- Clinical Trial Number: not applicable
- Ethics declaration: not applicable
- Consent to Publish declaration: not applicable
- Consent to Participate declaration: not applicable
- Competing Interests: The authors have no relevant financial or non-financial interests to disclose.
- Data Availability Statement: This study used the publicly available ASAP dataset developed by Crossley et al (Crossley, S., Baffour, P., King, J., Burleigh, L., Reade, W., & Demkin, M. (2024). Learning Agency Lab - Automated Essay Scoring 2.0. Kaggle). No new data were created.
- Authors Contributions: All authors contributed to the study conception and design. Material preparation, data collection and analysis were performed by FK, DSC, LS, DA, DRT, TK, and AA. The first draft of the manuscript was written by FK and all authors commented on previous versions of the manuscript. All authors read and approved the final manuscript.